%% file: acl_social.tex
\pdfoutput=1

\documentclass[11pt]{article}

\usepackage[final]{acl}

\usepackage{times}
\usepackage{latexsym}

\usepackage[T1]{fontenc}

\usepackage[utf8]{inputenc}

\usepackage{microtype}

\usepackage{inconsolata}

\usepackage{graphicx}

\usepackage{amsmath}
\usepackage{amsthm,enumitem }
\usepackage[ruled,vlined,linesnumbered]{algorithm2e}
\usepackage{booktabs}
\usepackage{balance}
\usepackage{colortbl}

\input{latex/math}

\title{Assistive Large Language Model Agents for \\ Socially-Aware Negotiation Dialogues\\\textcolor{red}{ \normalsize{WARNING: This paper may contain model outputs that are offensive, non-inclusive, or biased.}}}

\author{
 \textbf{Yuncheng Hua},
 \textbf{Lizhen Qu\footnotemark[2]},
 \textbf{Gholamreza Haffari}
\\
\\
 Department of Data Science \& AI, Monash University, Australia
\\
 \texttt{\{devin.hua, lizhen.qu, gholamreza.haffari\}@monash.edu}
}

\begin{document}
\maketitle

\renewcommand{\thefootnote}{\fnsymbol{footnote}}
\footnotetext[2]{Corresponding author.}

\begin{abstract}
\input{latex/section/sec-abs.tex}
\end{abstract}

\section{Introduction}
\label{Intro}
\input{latex/section/sec-intro.tex}

\section{Assistive Systems for Negotiations}
\label{Approach}
\input{latex/section/sec-approach.tex}

\section{Exemplars with High Value Impact}
\label{MCTS}
\input{latex/section/sec-mcts.tex}

\section{Experiments}
\label{Exp}
\input{latex/section/sec-exp.tex}

\section{Related Work}
\label{Related}
\input{latex/section/sec-related.tex}

\section{Conclusion}
\input{latex/section/sec-conclusion.tex}

\section*{Limitations}
\input{latex/section/sec-limitation.tex}

\section*{Ethics Statement}
\input{latex/section/sec-ethics.tex}

\section*{Acknowledgments}
\input{latex/section/sec-acknow}

\clearpage
\balance
\bibliography{acl_social}

\clearpage
\appendix
\section{Appendix}
\label{sec:appendix}
\input{latex/section/sec-appendix.tex}


\end{document}

%% file: latex/math.tex
\usepackage{amsthm,amsmath,amsfonts,bm,xspace}
\usepackage{upgreek}
\usepackage{color}

\newcommand{\comment}[1]{}

\newcommand{\chatgpt}{GPT 3.5\xspace}
\newcommand{\gptf}{GPT 4\xspace}

\newcommand{\infovalue}{ValueImpact}
\newcommand{\coin}{\textsf{coin}}
\newcommand{\rem}{\textsf{remediation}}

\def\eqref#1{(\ref{#1})}

\def\1{\bm{1}}

\DeclareMathAlphabet{\mathsfit}{\encodingdefault}{\sfdefault}{m}{sl}
\SetMathAlphabet{\mathsfit}{bold}{\encodingdefault}{\sfdefault}{bx}{n}



%% file: latex/section/sec-abs.tex
We develop assistive agents based on Large Language Models (LLMs) that aid interlocutors in business negotiations.
Specifically, we simulate business negotiations by letting two LLM-based agents engage in role play. A third LLM acts as a remediator agent to rewrite utterances violating norms for improving negotiation outcomes.
%
We introduce a simple tuning-free and label-free In-Context Learning (ICL) method to identify high-quality ICL exemplars for the remediator, where we propose a novel select criteria, called \textit{value impact}, to measure the quality of the negotiation outcomes. We provide rich empirical evidence to demonstrate its effectiveness in negotiations across three different negotiation topics. 
We have released our source code and the generated dataset at: {\small\textsf{\url{https://github.com/tk1363704/SADAS}}}.
%

%% file: latex/section/sec-intro.tex

There is a growing interest to build conversational agents with social intelligence, aiming to assist humans to achieve both task and social goals~\citep{gweon2023socially,wang2024sotopia}. 
%
Compared to task-oriented goals, such as booking a flight, the subjective nature of social goals, e.g. rapport building, makes them more challenging to model and quantify, especially when they often require social interactions. 
Machine social intelligence necessitates virtual agents to demonstrate human-like social behaviors and handle intricate social tasks like cooperation and negotiation~\cite{li2023metaagents,DBLP:journals/corr/abs-2310-02124}.

\begin{figure}[!t]
    \centering
    \includegraphics[width=\columnwidth]{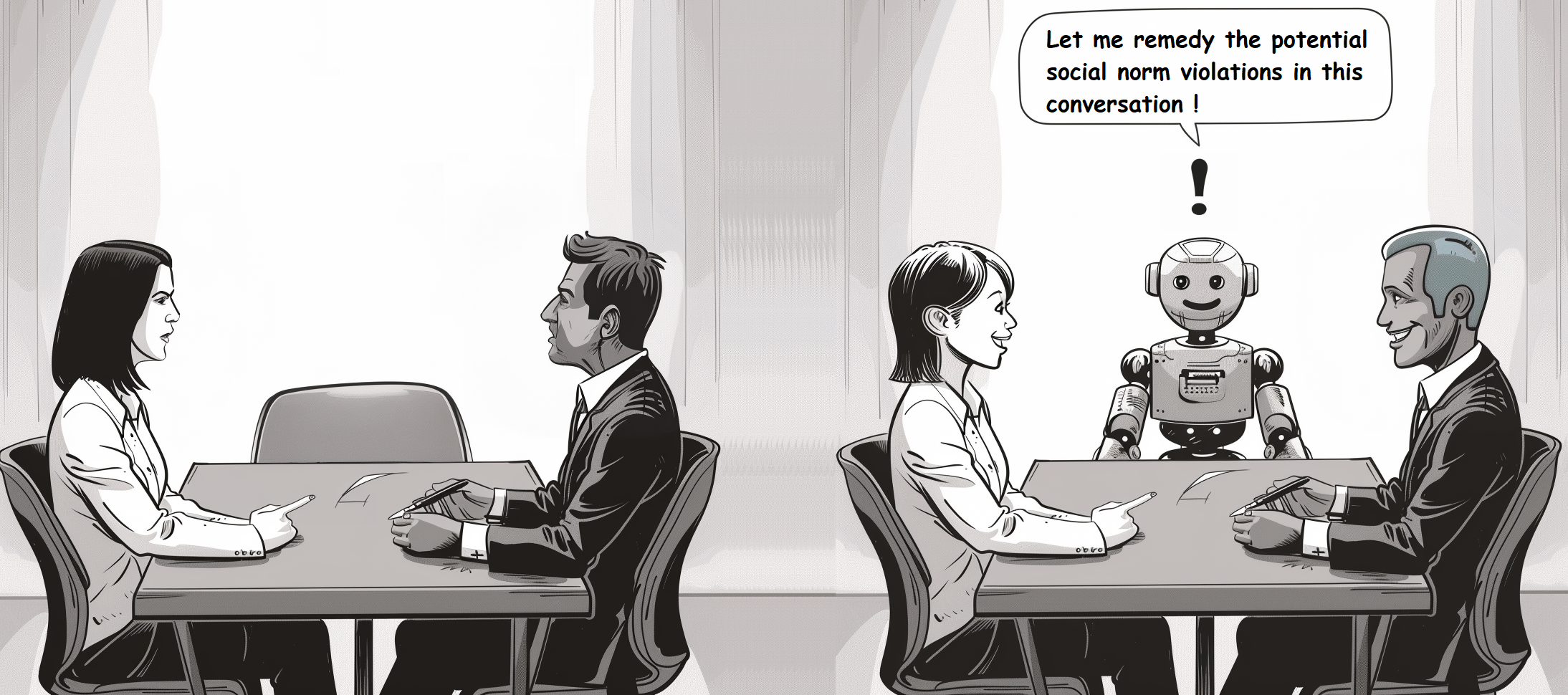}
    \caption{An illustration of our social agent(s). We expect to apply the remediator to real-world negotiations where the remediator can rectify potential social norm violations in the dialogue, thereby reducing conflicts and misunderstandings caused by cultural differences. }
    \label{fig:illustration}
    \vspace{-15pt}
\end{figure}

Recent literature studies agents in simulated environments to explore their social skills~\cite{DBLP:journals/corr/abs-2401-04620,DBLP:journals/corr/abs-2312-15907,DBLP:journals/corr/abs-2309-17234,akyurek-etal-2023-rl4f,bakhtin2022human, fu2023improving,DBLP:journals/corr/abs-2401-04620} and task-oriented skills ~\cite{zhou2023sotopia, park2023generative, wang-etal-2023-chatgpt-defend, hua2023war, xu2023exploring, light2023avalonbench, wang2023avalon}. 
We are instead interested in agents that can intervene and enhance the interaction of other agents (see  Figure~\ref{fig:illustration}). 

In this paper, we investigate how effectively agents can \emph{aid} conversational partners in achieving their social goals and thereby improve negotiation outcomes.
We specifically focus on studying impact of social norms in business negotiations, since negotiation is an integral part of the daily life~\cite{bazerman1993negotiating,lewicki2011essentials}.
We develop a socially intelligent \emph{remediator} agent that \emph{intervenes} in social interactions. The agent generates remediation to correct inappropriate language elements that do not align with social norms. 
We quantify the benefits of remediation from both task-oriented and social goals, thereby empowering the agent to aid by addressing both aspects. 

We adopt an in-context learning (ICL) approach to enable non-trainable black-box models serve as the remediation agent, as opposed to the previous work based on fine-tuning ~\cite{bakhtin2022human, wang2024sotopia}.  
Our novel scoring function for ICL demonstration selection eliminates the necessity of ground-truth output labels~\cite{DBLP:conf/iclr/LinRLDSCB024}. 
Our ranking score, dubbed \textit{value impact}, quantifies both task-specific and social goals to evaluate how effectively the remediator can assist interlocutors in business negotiations, as well as to better differentiate between positive and negative ICL examples. 
Several works have considered the ICL demonstration selection problem. However, they either focus solely on classification tasks~\cite{DBLP:journals/corr/abs-2405-02501,wang2024large}, require additional costs to train a retriever to choose optimal demonstrations~\cite{DBLP:conf/eacl/WangYW24}, or rely on the generation probability of ground-truth answers to select demonstrations~\cite{DBLP:journals/corr/abs-2308-12032,DBLP:journals/corr/abs-2401-06301}. 
%
%
%
%
In this work, our contributions are,
\begin{itemize}
    \item We formulate the problem of assistive systems to help with social aspects of negotiation dialogues as a multi-agent problem.  We then propose a multi-agent social interaction environment to simulate negotiation dialogues with interventions, that involve social norm violations/adherence, using role-playing LLMs. 
    
    \item We introduce a simple tuning-free and label-free ICL method that effectively improves the social intelligence of an assistive agent based on LLMs using a few carefully selected examples from the past simulated interactions. This is achieved based on our novel ICL sample selection criteria, \emph{value impact}, that captures the value of interventions based on both social and negotiation outcomes.
    
    \item Through experiments, we demonstrate that the remediator, using our ICL example selection method, outperforms all baselines in enhancing negotiation outcomes and mitigating social norm violations. Compared to the best baseline model, our remediator achieves a maximum improvement of $4\%$ in negotiation success rate ($86\% \to 90\%$), a $1.5\%$ increase in deal price (630,479 $\to$ 640,154), and a $3\%$ enhancement in the achievement rate of social goals ($82\% \to 85\%$).   
\end{itemize}

\begin{figure*}
\centering
\includegraphics[width=1.0\textwidth]{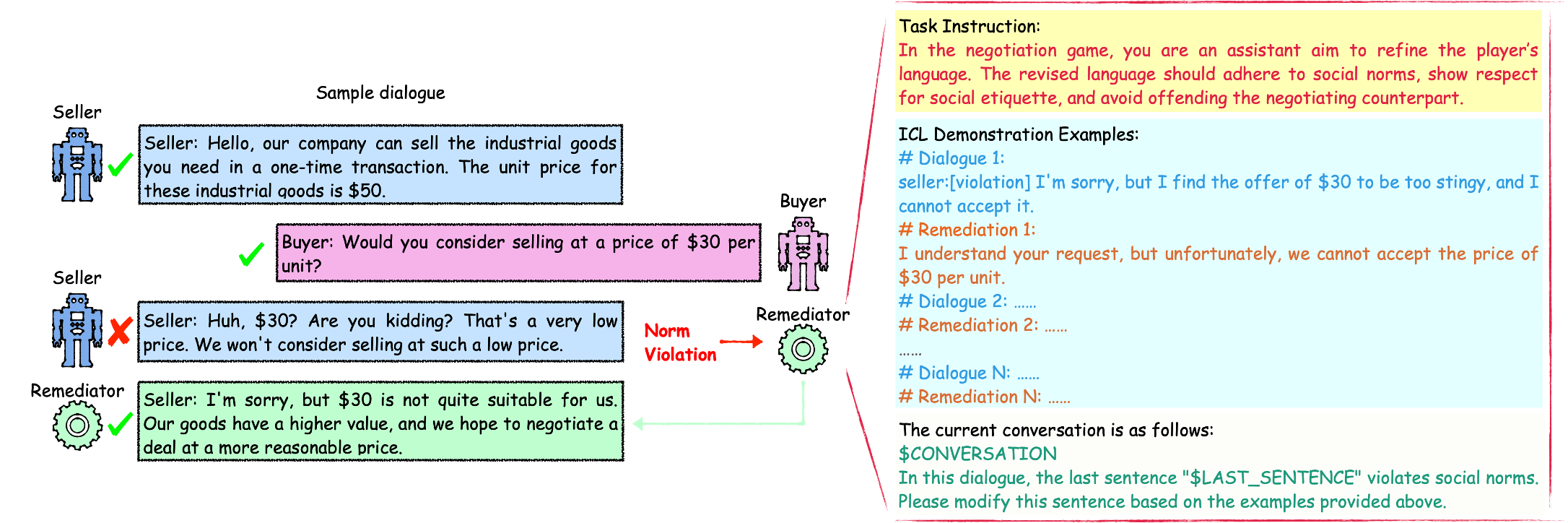}
\caption{A running example: In this conversational exchange between two interlocutors, an utterance from the seller breaches a social norm. Our remediator intervenes to rectify it by generating a remedial response.} \label{fig:example}
\end{figure*}







%% file: latex/section/sec-approach.tex
\subsection {Definition of Social Norm Violation}
\label{ssec:norm_definition}
From the perspective of computer science, particularly within computational social science and artificial intelligence, we provide a formal definition of social norm violation in this work.

A social norm violation is an action or behavior by an intelligent agent (human or artificial) that deviates from the established, implicit or explicit rules, expectations, or conventions governing acceptable conduct within a specific social context.

This deviation disrupts the behavioral equilibrium of the social system and is characterized by:
\begin{itemize}
    \item Divergence from expected behavior patterns as defined by the majority or authoritative entities within the system.
    
    \item Potential to elicit negative responses, sanctions, or corrective measures from other agents within the same social framework.
    
    \item Measurable impact on social dynamics, trust, cooperation, or cohesion among agents.

    \item Quantifiable deviation from formal models of normative behavior, often represented through game theory, multi-agent systems, or social network analysis.
\end{itemize}

We believe that understanding and predicting social norm violations are crucial in computational modeling and AI development to create systems that can effectively navigate complex social environments, make ethically aligned decisions, and maintain beneficial human–AI interactions.

\subsection {Problem Setting}
\label{ssec:problem_setting}
We simulate human negotiations with two LLM agents, assigning them respective roles and the goals they need to achieve. 
In the LLM-based negotiations, we mimic the real negotiations by injecting social norm violations in a controlled manner. 

In simulation, norm violations are viewed as a mapping to real situations, where, during a deadlock or intense negotiation, a negotiator might unintentionally use language that is overly aggressive, offensive, and violates social norms.

To mitigate the potential negative impact of such language on negotiation outcomes, a third-party LLM agent, the remediator, is introduced in this work to correct instances of norm violations.
The remediator aims to ensure that the language adheres to the social norms, and avoids offending the negotiating parties.
This in turn assists the parties in achieving their task goals and relationship goals, including building trust, deepening relationships, and establishing a stronger network between the conversational participants.
It is believed that the achievement of relationship goals will also impact the negotiation process, making it easier for the buyer and seller to reach an agreement when there is a mutual understanding and trust.

The remediation generated by the remediator is used to rewrite sentences involving norm violations and conveyed to the counterpart party to facilitate smooth dialogue.
To focus on the functional study of the remediator, we consistently have the seller generate dialogues that may involve norm violations, while the buyer engages in normal conversation. 
So, the remediator only corrects the language used by the seller.

We employ two role-playing language agents, i.e., the buyer and seller, and an remediator to simulate the realistic human negotiations with socio-cultural norm violation. The details of the implementation and the algorithm for the LLM-based simulation can be checked in Appendix~\ref{appendix_simulation}.

\subsection{LLM-based Assistive Agents}
\label{baselines}
The generative LLMs obtained through extensive pre-training inherently possess the ability for semantic understanding and task insturction following. Leveraging past experiences learned from the corpus, these models can, in a zero-shot learning manner, to some extent address a new downstream task. However, the data distribution of the training corpus for such untuned models may differ from that of the downstream task, leading to issues of distributional bias. 

Consequently, when untuned models handle downstream tasks, the following problems may arise: 1. The model may not strictly adhere to task instructions, generating redundant information beyond task requirements; 2. The generated content may be inconsistent with the preferences of the downstream task. Since untuned models have not undergone sufficient training in the downstream task and thus lack specific knowledge about it, they often struggle to provide effective assistance in conversations. Therefore, we first introduce (or design) baseline methods with different architectures based on the negotiation task (the details can be viewed in Appendix~\ref{appendix_baselines}). Subsequently, we present our memory-augmented ICL model.

%% file: latex/section/sec-mcts.tex
\subsection{Exemplar Filtering using Value Impact}
\label{ssec:individual_filter}
A core challenge in the design of our Assistant is that we aim to achieve high-quality norm remediation with low or zero training costs\footnote{Especially for black-box models like the ChatGPT, GPT-4o, and Claude series LLMs that cannot be trained.} and minimal inference time.
Many related works have demonstrated that a small number of constant stylistic ICL examples can enhance the alignment of LLMs with downstream tasks~\cite{DBLP:conf/iclr/LinRLDSCB024}.
Therefore, we search for approximately optimal ICL examples offline and encapsulate them into prompts to determine the agent's policy.
This approach has two main advantages: first, by performing calculations offline, the LLM avoids the time needed for online learning, thus reducing inference time. Second, ICL learning does not require training, which reduces training costs.
We denote the agent's policy by $\pi_{\theta}$ where $\theta$ is the set of chosen training exemplars.  
In this section, we present a method for selecting such crucial exemplars in order to characterise a near optimal policy $\pi_{\theta^*}$.

Let $d=(h_{<t}, x_t, y_t, h_{>t})$ be an annotated dialogue between the buyer and the seller, where $x_t$ is the $t$-th dialogue turn and $y_t$ is its \emph{silver} groundtruth remediation (annotated by a zero-shot GPT 3.5 that none of the ICL exemplars is provided), $h_{<t}$ denotes the conversation history from the start up to the turn $t$, and $h_{>t}$ denotes the continuation of the conversation to the end. Let $R(d)$ denotes the \emph{final reward/outcome} of the conversation, encompassing various factors such as whether a deal was reached, the agreed  price, the change in the quality of the business relationship due to this dialogue, and the change in the quality of the trust after this dialogue. 
%


Let $D$ be the  dialogue dataset annotated with the \emph{silver} remediations. For each dialogue $d \in D$ and a turn $t$ with norm violation, we consider $z=(h_{<t}, x_t, y_t)$ to be a candidate exemplar that can be included in the agent's memory. 


\paragraph{Value of an remediation} Consider an exemplar $z=(h_{<t}, x_t, y_t)$ extracted from a dialogue $d=(h_{<t}, x_t, y_t, h_{>t})$.   
We let the two role-playing LLM agents randomly synthesize a new business negotiation task and begin their dialogue until a remediation point $x_s$ is reached.
For remedying $x_s$, we first feed a prompt without any ICL examples to the remediator agent, allowing it to generate a \emph{silver} remediation $y_s$ in a zero-shot learning setting. Subsequently, we pack $z$ as an ICL example into the task instruction and prompt the remediator to generate remediation $y'_s$ in a one-shot learning setting.
We define the \emph{value} of $y'_s$ wrt the \emph{silver} remediation $y_s$ for an remediation point $x_s$ as,
\begin{eqnarray}
\label{eqn:val}
\fontsize{9pt}{9pt}\selectfont
\begin{aligned}
 V_z(y'_s) &:= E_{p_{\textrm{sim}}({h'}_{>s}|y'_s,x_s, h_{<s})} \cdot R(h_{<s}, x_s, y'_s, {h'}_{>s}) \\
  & - E_{p_{\textrm{sim}}({h}_{>s}|y_s,x_s, h_{<s})} \cdot R(h_{<s}, x_s, y_s, h_{>s})
\end{aligned}
\end{eqnarray}
where $p_{\textrm{sim}}(h_{>s})$ is the distribution over possible completions of the dialogue, following the remediation and the conversation history. 
We can sample from $p_{\textrm{sim}}(h_{>s})$ using simulation Algorithm \ref{simulation}. 
%
Some remarks are in order: (1) A complete trajectory $(h'_{<s},x_s,y',h'_{>s})$ is composed of actions of three agents, i.e. the assistive remediator agent as well as the role-playing LLM agents for the buyer and seller.  
We are mainly interested in the value of information for the actions taken by the remediator agent. 
(2) Due to high simulation cost, in the experiments: (i)  we sample $h'_{>s}$ and $h_{>s}$ once to \emph{estimate} the value of information using eqn \ref{eqn:val}, and (ii) 
we allow only one remediation point  in a simulated dialogue. 

We designed a heuristic-based reward calculation formula. We use a GPT-3.5-based evaluator agent to assess the dialogue status and extract the transaction price, $v_{price}$, which is then normalized to the [0,1] range based on the price interval. We use $b_{deal}$ to indicate whether a deal was reached at the end of the dialogue. If the transaction is completed, $b_{deal} = 1$; if the transaction is not completed, $b_{deal} = -1$. Additionally, the evaluator assesses changes in the social relationship status of the dialogue participants before and after the negotiation. We use $\delta_{trust}$ and $\delta_{bus}$ to represent whether trust and business relationships between the participants have strengthened post-negotiation, respectively. If trust or business relationships is deepened, the value is 1; if there is no change, the value is 0; if it becomes worse, the value is -1. Thus, we propose the following formula to quantify the social goal and task-oriented goal:
\begin{eqnarray}
\label{eqn:social}
\fontsize{9pt}{9pt}\selectfont
\begin{aligned}
 R(d) = \alpha \cdot v\_{price} + \beta \cdot b\_{deal} \\ + \gamma \cdot \delta\_{trust} + \epsilon \cdot \delta\_{bus}
\end{aligned}
\end{eqnarray}
\vspace{-20pt}

%
%
%
%

\paragraph{Value Impact of exemplar(s)} We use the role-playing agents to synthesize dialogues $D_s$, and define the \emph{value impact} of exemplars as the \emph{values} that they produce when used in the ICL policy to remediate $D_s$. 
Consider one ICL exemplars $\tilde{z}$, we define the value impact of it as: 

\begin{eqnarray}
\label{eqn:impact}
\fontsize{9pt}{9pt}\selectfont
\begin{aligned}
  V^{\textrm{impact}}_{Z=\{\tilde{z}\}} := \sum_{x_s \in D_s} V_{\tilde{z} }(\pi_{Z=\{\tilde{z}\}}({x}_s, {h}_{<s}))/|D_s|
\end{aligned}
\end{eqnarray}
where $\tilde{z} = (\tilde{h}_{<t}, \tilde{x}_t, \tilde{y}_t)$ is an exemplar,  $\pi_{Z=\{\tilde{z}\}}({x}_s, {h}_{<s})$ is the remediation generated by the policy $\pi_{Z=\{\tilde{z}\}}$, and $|D_s|$ is the size of the synthetic dialogues. The policy $\pi_Z$ is built with an LLM using the three-part prompt structure of Figure \ref{fig:example} and one ICL exemplar set $Z=\{z\}$.
It should be noted that we can include multiple ICL examples in $Z$, transitioning the remediator from a one-shot learning setting to a few-shot learning setting.


As the policy is characterised by the examples included in the  memory prompt, the problem of optimising the policy boils down to choosing the optimal subset of examples $Z^*$ from $D$ to include in the prompt to maximise the value impact,
\begin{eqnarray}
   Z^* = \arg\max_{Z \subseteq D} V^{\textrm{impact}}_Z 
\end{eqnarray}
%
We explain our optimisation algorithm for choosing such near optimal examples as follows. 

\paragraph{Individual exemplar Filtering} In the first step, we search for individual norm violation examples with high \emph{value impact}. 
We rank the candidate examples in $D$ according to their individual  value impact $V^{\textrm{impact}}_{\{z\}}$. 
%
Computing the individual value impacts based on eqn \ref{eqn:impact}  can be time consuming due to several reasons: (i) the large number of candidate examples in $D$,  (ii) the high computation needed to compute the exact expectation for the value $V$ in eqn \ref{eqn:val}, and (iii) the high computation needed for computing the expectation over a large example set $D_s$ in eqn \ref{eqn:impact}. We thus resort to approximations: (i) we only consider a subset of $|S'| << |D|$ as candidate examples for ranking chosen randmoly from $D$ and (ii) we approximate the value impact in eqn \ref{eqn:impact} based on a small sized $D_s$. We then rank the candidate examples in $S'$ according to their estimated value impact for the next stage of optimisation. 


\begin{figure}[!t]
    \centering
    \includegraphics[width=\columnwidth]{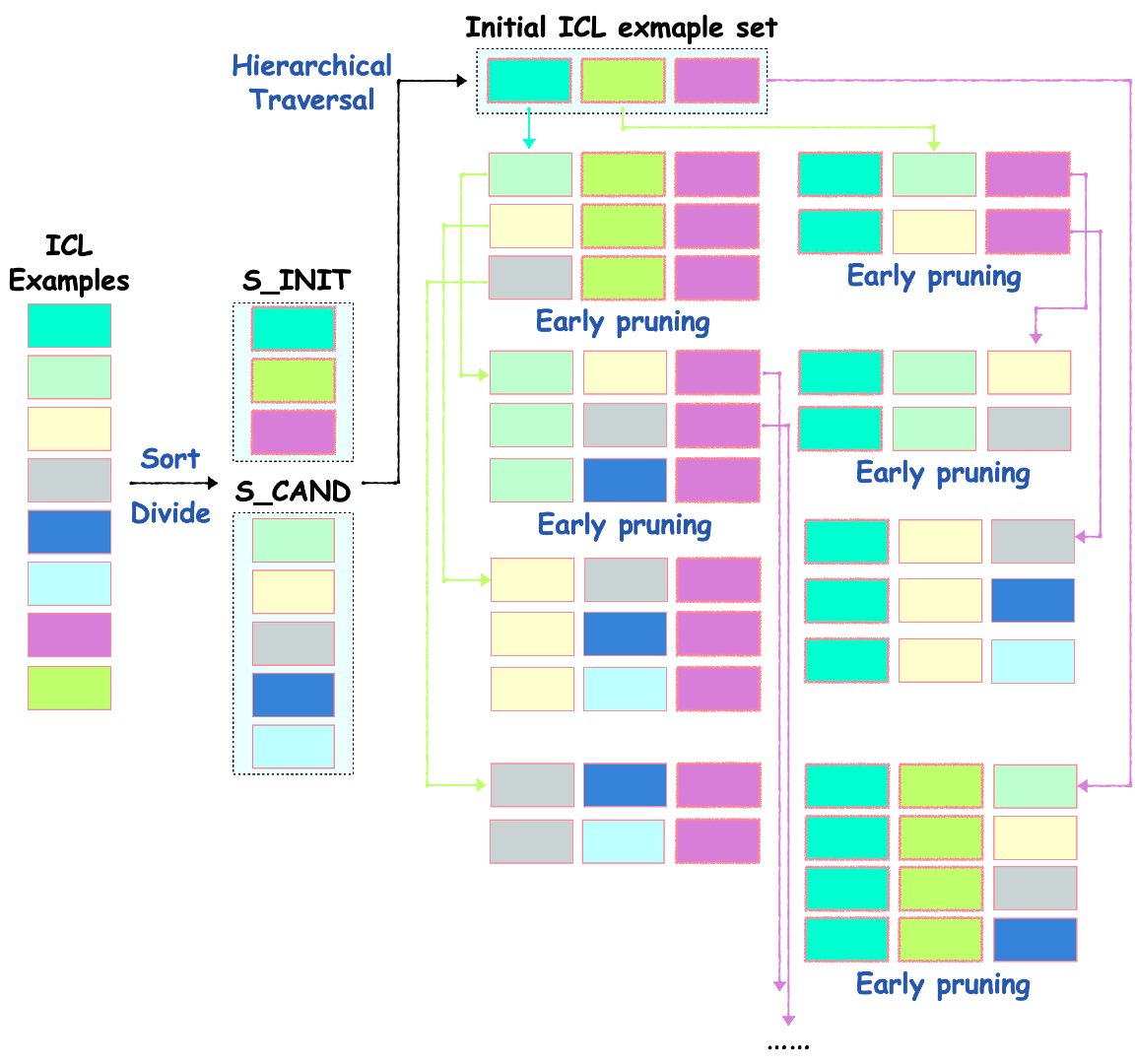}
    \caption{An illustration of using Hierarchical traversal with early pruning to search for the optimal exemplars.}
    \label{fig:DFS}
    \vspace{-15pt}
\end{figure}

\subsection{Selection of a Set of ICL Demonstrations}
\label{ssec:dfs}
\paragraph{Search for Optimal exemplars.} The combination of multiple ICL exemplars often provides more assistance to the model in tackling tasks, compared to a single ICL exemplar.
%
After completing the first step, we now need to find an approximate optimal ICL example \emph{set} to build an effective policy. 

As shown in Figure~\ref{fig:DFS}, we sorted all exemplars in $S'$ in descending order according to their value impact and selected the top-$n$ exemplars with the highest individual value impact to form an ICL exemplar set, i.e., $S_{INIT}$. The remaining ICL examples in $S'$ (also sorted in descending order of value impact) are considered as the candidate ICL exemplar pool, i.e., $S_{CAND}$. We regard the $S_{INIT}$ as the initial ICL example \emph{set} $S_{ICL}$.

Our goal is to combine ICL examples from $S_{INIT}$ and $S_{CAND}$ sequentially using a hierarchical traversal algorithm. This approach aims to search for different combinations of ICL examples and select the one combination with the highest value impact, representing the approximately optimal ICL example set we seek to explore.
The computational complexity of hierarchical traversal is $\mathcal{O}(|S'| \cdot |D_s| \cdot |S_{INIT}| \cdot |S_{CAND}|)$.
It becomes evident that as the size of $|S|$ increases, the search space expands significantly, and the search approximates an NP-hard problem.
%
%
During our empirical study, we found that the value impact of a single ICL exemplar holds certain indicative significance. Specifically, if an ICL exemplar has a higher value impact, then if this exemplar remains in the ICL example \emph{set}, the overall value impact of the set will also be correspondingly higher. Therefore, the probability of this exemplar staying in the final ICL example \emph{set} is higher compared to exemplars with lower value impact. Based on this discovery, we designed a heuristic rule for early pruning during hierarchical traversal, which will be detailed in the following sections.

We initialize an empty queue $q$ and enqueue $S_{INIT}$, starting the hierarchical traversal. 
In each iteration, we dequeue the elements of the current level from $q$, with each element being a combination of ICL exemplars $S'_{ICL}$. For an ICL exemplar $a$ originally in $S_{INIT}$ within $S'_{ICL}$, we sequentially select an exemplar $b$ from $S_{CAND}$ based on its sorted order and replace $a$ in $S'_{ICL}$ with $b$ to form a new $S''_{ICL}$, making it a child node of $S'_{ICL}$. 
We then calculate the value impact change of this new $S''_{ICL}$ and enqueue it into $q$ as a child node to be visited in the next level.
The value impact change is: $\Delta := V^{\textrm{impact}}_{S''_{ICL}} - V^{\textrm{impact}}_{S'_{ICL}}$.

Notably, if we observe that the $\Delta$ of $S''_{ICL}$ is negative for $M$ consecutive replacements, we conclude that it is unnecessary to continue replacing $a$ with further lower-ranked $b$ exemplars from $S_{CAND}$. 
Consequently, we terminate the exploration of the current branch and do not enqueue further child nodes of $S'_{ICL}$ (by replacing $a$) into $q$, thereby completing early pruning.
When all elements in the queue $q$ have been dequeued and visited, the hierarchical traversal ends. 
At the end, we select the ICL example \emph{set} with the highest value impact as our final solution, therefore, we obtain $\pi^* := \pi_{S^*_{ICL}}$, which is considered as an approximately locally optimal policy for remediation.

The details of the algorithm can be found in Appendix~\ref{appendix_hierarchical}.

%% file: latex/section/sec-exp.tex
\subsection{Experimental Settings}
\paragraph{Dialogue generation.} We consider three topics: housing price, product sale, and salary negotiation for bilateral negotiations. For the negotiator agents, we design topic-specific prompts for each role to inform the topic, task-related goals and relational goals, as shown in Table~\ref{tab:seller_violate_prompt},~\ref{tab:seller_no_violate_prompt}, and~\ref{tab:buyer_prompt} in Appendix~\ref{appendix_prompt}.
To minimize the impact of confounders on the generated dialogues, we use same definitions for the relational goals across all topics. The prompts vary in terms of topics, task-related goals, and the description of roles. For example, in salary negotiation, the task-orient goal of the job seeker is to obtain the highest possible salary, while the goal of the employer is to match the job seeker's work abilities with appropriate salary. 

To enhance the diversity of the simulated dialogues, we provided substantial freedom in the dialogue generation process. 
During the simulated negotiations, we only hard-coded the first two lines of each dialogue (see the prompts used in Table~\ref{tab:seller_violate_prompt}, \ref{tab:seller_no_violate_prompt} and \ref{tab:buyer_prompt}), allowing the agents to autonomously continue the conversation. 
We manually intervened only in negotiation processes that entered deadlocks or repetitive loops, thereby granting maximum freedom to all other dialogues. 
This approach enabled the negotiator agents to converse as freely as possible, thus increasing the structural complexity and content diversity of the dialogues.

To add utterances with norm violations into the dialogues, we set $p_c$ to 0.4,  as described in Appendix~\ref{appendix_simulation}. The resulting proportion of turns with violations in each dialogue is approximately 44.36\%. In each dialogue, we assume that only one agent is not aware of social norms to produce those violations, which is the seller for housing price and product sale, and the job seeker for salary negotiation. As a result, we are able to use the metrics introduced below to consistently assess the quality of the remediation models, the higher the better.

For the negotiator agents, we adopt \chatgpt\footnote{\url{https://openai.com/}} as the LLM to produce dialogues in Chinese. Chinese is chosen because there is a high probability that \chatgpt does not produce English dialogues with norm violations due to violations of the OpenAI policies.
It is important to note that the focus of this research is on analyzing cross-linguistic social norm violations (actions that are considered violating social norms in different linguistic contexts), rather than culture-specific social norms (e.g., behaviors that are considered adhering to social norms in Western culture but are seen as norm violations in Eastern culture). 

Following Algorithm~\ref{simulation}, we generate 100 dialogues per topic as the test set, while generating 333 dialogues per topic for training. The training set also serves as the pool for ICL exemplars.
We have released our generated and collected dataset at {\small\textsf{\url{https://github.com/tk1363704/SADAS}}}.

\paragraph{Remediation Baseline Models.} As this work focuses on understanding the impact of remediations, we apply the remediation agents directly to each turn marked with violations to produce remediations without employing any detection models.  

We have the following baselines as described in Section~\ref{baselines} and~\ref{MCTS}: \textbf{Prompt-based LLM}: use a task instruction (without ICL examples) prompt LLM to remedy norm violations. \textbf{Vanilla ICL-based LLM}: randomly selecting $K$ ICL demonstration examples from $D$ to compose prompt; \textbf{RL-based LLM}: summarizing the past dialogues and remediation and incorporating the summary into the content of randomly selected ICL demonstration examples. \textbf{SFT-based LLM}: employing $D$ to supervised finetune the Atom-7B-Chat\footnote{\url{https://github.com/FlagAlpha/Llama2-Chinese}} (a Llama2-7B model that is finetuned using a substantial amount of Chinese corpus). We equip Atom with Low-rank adaptation (LoRA) for finetuning. \textbf{Retrieval-augmented ICL-based LLM}: to retrieve top-K examples in $D$ that are the most similar to the current query dialogue as the demonstration examples. \textbf{\infovalue \  ICL-based LLM}: find the ICL demonstration example set that has the highest Impact Value. The example set is consisted of $K$ examples, and we set $M=2$. In all ICL-related methods, $K$ is set as 8. Since \chatgpt is an untrainable model, we implement all the prompt-based and ICL-based methods using both \chatgpt and Atom-7B-Chat, but only implement the SFT-based method using Atom. We can refer to Appendix~\ref{appendix_baselines} to check the details of the implementation of the baseline models. 

In eqn~\ref{eqn:social}, we set $\alpha=0.7$, $\beta=0.1$, $\gamma=0.1$, and $\epsilon=0.1$, respectively.

\input{data/table/big_results}

\subsection{Metrics}
We evaluate the remediation models based on negotiation outcomes from four perspectives: \textit{success rate}, \textit{deal value}, \textit{trust improvement}, and \textit{relation enhancement}. The former two are calculated by rules, while the latter two are evaluated by using \gptf with the designated prompts outlined in Table~\ref{tab:trust_prompt} and~\ref{tab:business_prompt} in Appendix~\ref{appendix_prompt}. We consider evaluating negotiation outcomes because our Algorithm~\ref{simulation} for dialogue generation view remediation measures as interventions so that different remediations lead to different flows of conversations. As a result, there are no groundtruth responses to compare with because there are exponentially many possibilities that a conversation can take.

We define four metrics to evaluate the outcome of the negotiations, including: (1) Success Rate (Suc): the percentage of negotiations that end up with successful deals. (2) Deal Value (Deal(\$)): the agreed final deal price after an negotiation averaged across all conversations. (3) Trust Improvement (Trust): the ratio of the negotiations that an agent obtain a \textit{higher} trust from the counterpart than that at the begin of conversations. (4) Relation Enhancement (Rel):  the percentages of the negotiations that an agent has \textit{better} relation with the other party at the end of negotiations. The details of the metrics can be viewed in Appendix~\ref{appendix_metrics}.  



\input{data/table/ablation_study}

\input{data/table/human_evaluation_result}

\subsection{Results and Analysis}
\label{ssec:big_result}


We conduct experiments to show the effectiveness of the remediation agent using our proposed method, in comparison with competitive baselines. 
From Table~\ref{tab:result} we can see that norm violations consistently harm the outcomes of negotiations w.r.t. all four perspectives if no remediation applies. This aligns well with the Expectation Violation theory in social science~\cite{levine2000norms}.

Remediation effectively improves the negotiation outcomes for almost all models w.r.t. all metrics across all topics, except for very few cases, e.g. the success rate of the prompt-based model using Atom-7B-Chat for ``product sale'' is 2\% lower than that without any remediation. Those LLM-based agents can indeed help negotiation agents achieve their relational goals, and further improve negotiation success rates and deal values, regardless if the LLMs are fine-tuned or not.

Our approach based on \chatgpt, denoted as \infovalue \  ICL in Table \ref{tab:result}, consistently outperform all baselines in terms of all metrics. When the LLM is switched to Atom-7B-Chat, there are slight performance drops in all metrics, which shows the importance of the ability of LLMs to understand ICL examples and prompts. Despite that, our approach with this open-source LLM achieves still superior performance than the baselines using the same LLM in most of the cases.

The most relevant method to our approach is Retrieval ICL, which identifies $K$ nearest neighbours as ICL examples. As our method outperforms Retrieval ICL in almost all cases, the ICL example set using our approach is indeed better than the widely $K$ nearest examples selected on the fly. Furthermore, when we compare the ICL examples used in Retrieval ICL with those using our approach, we find that the overlap rate is approximately 40\%. It is evident that the best ICL examples are not necessarily the widely used $K$ nearest neighbours.


Additionally, we observe a consistent trend across the three topics, combining four metrics: except for a few cases, the zero-shot prompt-based LLM implemented remediator generally performs lower than the SFT LLM. The SFT LLM's performance is inferior to ICL-based LLMs. Within the ICL-based LLM family, the Vanilla ICL model, derived from random ICL examples, exhibits the poorest performance. The RLNL, which incorporates NL feedback, performs better. The nearest neighbor ICL examples obtained through similarity retrieval show intermediate performance. Notably, the \infovalue \  ICL proposed in this paper exhibits the best performance.


The reason for the inferior performance of SFT LLM compared to ICL-based LLM is the relatively small size of the pseudo-gold annotation set $D$ (approximately 1000 instances). This limited quantity hinders the effective optimization of parameters, preventing the model from fully learning task-relevant knowledge. Simultaneously, it is likely that due to this reason, and because Llama2 has limited support for Chinese, methods implemented based on Atom generally perform weaker than their counterparts implemented based on \chatgpt.

\paragraph{Ablation Study}
In our ablation study, as shown in Table~\ref{tab:ablation}, we experimentally evaluated the impact of Value Impact, topic diversity, hierarchical traversal, and the M-value in hierarchical traversal on overall model performance. 
We obtained the following key conclusions: (1) Value Impact plays a crucial role in identifying the optimal ICL examples. (2) The higher the diversity of ICL examples, the better the results. (3) Compared to simply combining individual ICL exemplars with the highest Value Impact, hierarchical traversal retrieves better combinations of ICL demonstrations. (4) The M-value represents the search space of hierarchical traversal. When the M-value is too small, retrieval performance is poor; when the M-value is increased, it does not significantly improve the quality of ICL demonstrations and results in a lot of ineffective search computations. Therefore, M=2 is our most cost-effective choice. For additional details on the Ablation study and related experimental results table, please refer to the Appendix~\ref{appendix_ablation}.

\paragraph{Human Evaluation}
As shown in Table~\ref{tab:human_eval}, we employed three annotators to conduct human evaluations on four baseline models and our own model across dialogues in three different topics. 
We evaluated two aspects of the conversations: (1) whether the dialogues were fluent and logically realistic after remediation (Dialogue column in Table~\ref{tab:human_eval}), and (2) whether the remediation effectively corrected norm violations, helped negotiators achieve better outcomes, and fostered positive social relationships with counterparts (Social Norm Remediation column in Table~\ref{tab:human_eval}).

According to the results in Table~\ref{tab:human_eval}, similar to the findings in Table 1, the models performed from best to worst as follows: ValueImpact ICL > Retrieval ICL > RLNL > Vanilla ICL > PROMPT. 
Our method, ValueImpact ICL, scored highest in overall dialogue quality assessment, effectiveness of remediation, and assistance provided. 
RLNL, by transmitting natural language feedback generated by a LLM agent to other LLM agents, enabling other agents to learn how to negotiate using strategies, thus producing more natural and logically coherent dialogues than Retrieval ICL.
However, in terms of the quality of the remediations, Retrieval ICL outperforms RLNL in helping negotiators achieve better transaction outcomes.

Using the Plausibility and Coherence metrics, our annotators evaluated the realism of the simulated dialogues.
The higher the scores, the more closely the simulated dialogues align with real-world scenarios, making it more likely that the remediator, trained through simulated scenarios, can assist in real-world negotiations.
Compared to baseline models, our scores were the highest, indicating that our scenarios are the most realistic for negotiation.
For more details and the design of metrics within human evaluation, please refer to the Appendix~\ref{appendix_human_eval}.

\paragraph{Computation Cost}


 
We need to consider the question: how much computational time does our method require? The answer is that, since we use a single ICL example set for all tasks, the inference complexity remains $\mathcal{O}(1)$.
Additionally, would increasing the training samples makes SFT outperform the ICL method? The answer is that, increasing the data does not significantly improve the performance of SFT.
The details of the experiments and discussions regarding these two questions can be found in Appendix~\ref{appendix_computation}.

%% file: data/table/big_results.tex

\begin{table*}[htbp]
\centering
\fontsize{8pt}{10pt}\selectfont
\begin{tabular}{c|cccc|cccc|cccc}
\hline
Topic → & \multicolumn{4}{c|}{Product Sale}          & \multicolumn{4}{c|}{Housing Price}                    & \multicolumn{4}{c}{Salary Negotiation}               \\ \hline
Method ↓       & Suc.      & Deal (\$)     & Trust & Rel.  & Suc.     & Deal (\$)     & Trust     & Rel.  & Suc.      & Deal (\$)   & Trust  & Rel.  \\  \hline
Without Viol.  & 90\%      & 42.13         & 78\%  & 84\%  & 78\%     & 646125        & 74\%      & 76\%  & 90\%      & 3487.5      & 74\%   & 80\%  \\
Viol No-Remed. & 74\%      & 38.14         & 66\%  & 70\%  & 60\%     & 594867        & 64\%      & 66\%  & 80\%      & 3371.5      & 68\%   & 70\%  \\ \hline
\multicolumn{13}{c}{With Violation (\chatgpt)}    \\ \hline
PROMPT          & 76\%      & 40.66         & 72\%  & 78\%  & 66\%     & 617580        & 66\%      & 68\%  & 84\%      & 3393.0      & 70\%   & 72\%  \\
Vanilla ICL    & 78\%      & 41.08         & 74\%  & 78\%  & 68\%     & 620176        & 70\%      & 70\%  & 86\%      & 3457.7      & 70\%   & 74\%  \\
RLNL           & 77\%      & 41.18         & 74\%  & 80\%  & 70\%     & 622479        & 70\%      & 72\%  & 84\%      & 3450.6      & 70\%   & 72\%  \\
Retrieval ICL  & 80\%      & 41.57         & 76\%  & 82\%  & \textbf{76\%} & 630479   & 72\%      & 74\%  & 86\%      & 3484.5      & 74\%   & \textbf{76\%}  \\
\infovalue \ ICL& \textbf{82\%} & \textbf{42.20} &\textbf{78\%} & \textbf{85\%} &\textbf{76\%} &\textbf{640154} &\textbf{75\%} &\textbf{76\%} &\textbf{90\%} & \textbf{3506.0} & \textbf{76\%} & 75\% \\ \hline
\multicolumn{13}{c}{With Violation (Atom-7B-Chat)} \\ \hline
PROMPT         & 72\%       & 39.24         & 70\%  & 72\%  & 62\%     & 608977        & 64\%      & 65\%  & 81\%      & 3409.4      & 70\%   & 70\%  \\
SFT           & 75\%       & 40.70         & \textbf{74}\%  & 78\%  & 66\%     & 618471        & 68\%      & 68\%  & 84\%      & 3405.5      & 70\%   & 72\%  \\
Vanilla ICL   & 76\%       & 41.10         & 72\%  & 77\%  & 66\%     & 619902        & 69\%      & 67\%  & 84\%      & 3410.7      & 71\%   & 71\%  \\
RLNL          & 76\%       & 41.23         & 72\%  & 76\%  & 68\%     & 619875        & 68\%      & 70\%  & 83\%      & 3408.3      & 71\%   & 72\%  \\
Retrieval ICL & 77\%       & 41.13         & 72\%  & 76\%  & 70\%     & 620974        & 69\%      & \textbf{71}\%  & 85\%      & 3455.8      & 72\%   & 73\%  \\
\infovalue \ ICL    & \textbf{79}\%       & \textbf{41.80}         & 73\%  & \textbf{79}\%  & \textbf{71}\%     & \textbf{627834}        & \textbf{71}\%      & 70\%  & \textbf{86}\%      & \textbf{3460.6}      & \textbf{73}\%   & \textbf{74}\%     \\ \hline      
\end{tabular}
\caption{The evaluation of remediation models on negotiation outcomes. The row `Without Viol.' denotes the setting that no norm violations occur in any conversations, while the row `Viol No-Remed.' refers to the negotiations with violations but no remediation models are applied. The remediation models below 'With Violation (\chatgpt)' are based on \chatgpt, while the models below 'With Violation (Atom-7B-Chat)' are the ones using Atom-7B-Chat.}
\label{tab:result}
\end{table*}

%% file: data/table/ablation_study.tex
\begin{table}[t]
    \centering
    \fontsize{8pt}{10pt}\selectfont
    \begin{tabular}{c|cccc}
    \hline
            Product Sale    &  Suc.  & Deal (\$) & Trust & Rel.\\ \hline  
         \multicolumn{5}{c}{Standard (\chatgpt)} \\ \hline
         \ \ \ Vanilla ICL    &  78\%     & 41.08 &   74\%    &     78\%    \\
         \ \ \ Retrieval ICL        & 80\%  & 41.57 &   76\%    &    82\%  \\
         \ \ \  \infovalue \ ICL      & 82\%    & 42.20 &   78\%    &     85\%   \\ \hline 
          \multicolumn{5}{c}{Ablation (\chatgpt)}  \\ \hline 
         \ \ \ Top \infovalue\  ICL        &   81\% &  41.78 &  76\% &  83\% \\
         \ \ \ Topic retrieval ICL  &  79\%  &  41.33 & 76\%  &  81\% \\
         \ \ \ Topic  \infovalue\  ICL      &  80\%  & 41.91 &  78\%  & 82\%  \\ 
         \ \ \ \infovalue \ ICL (M=5)      &  82\%  & 42.31 &  79\%  & 83\%  \\
         \ \ \ \infovalue \ ICL (M=1)      &  81\%  & 42.07 &  78\%  & 82\%  \\   \hline
    \end{tabular}
    \caption{The ablation study results.}
    \label{tab:ablation}
\vspace{-1em}
\end{table}

%% file: data/table/human_evaluation_result.tex
\begin{table*}[t]
\centering
\fontsize{8pt}{10pt}\selectfont
\begin{tabular}{c|cc|ccccc}
\hline
Target → & \multicolumn{2}{c|}{Dialogue}          & \multicolumn{5}{c}{Social Norm Remediation} \\ \hline
Method ↓            & Plau.     & Coher.    & Eff.  & Help Deal. (\%)           & Help Outcome. (\%)            & Trust (\%)              & Business Rel. (\%)  \\  \hline
PROMPT    & 2.18      & 2.27      & 2.17  & 66.1/ 23.2/ 10.7 & 58.9/ 23.2/ 17.9  & 33.9/ 12.5/ 53.6        & 71.4/ 17.9/ 10.7  \\
Vanilla ICL    & 2.20      & 2.30      & 2.25  & 67.9/ 21.4/ 10.7 & 60.7/ 23.2/ 16.1  & 35.7/ 10.7/ 53.6        & 75.0/ 14.3/ 10.7  \\
RLNL    & 2.35      & 2.62      & 2.35  & 69.6/ 17.9/ 12.5 & 71.4/ 12.5/ 16.1  & 42.8/ 5.4/ 51.8        & 80.4/ 10.7/ 8.9  \\
Retrieval ICL    & 2.33      & 2.58      & 2.37  & 73.7/ 15.8/ 10.5 & 68.4/ 15.8/ 15.8  & 42.1/ 5.3/ 52.6        & 78.9/ 10.5/ 10.5  \\ 
\infovalue \ ICL    & 2.49      & 2.68      & 2.43  & 79.5/ 9.0/ 11.5 & 77.0/ 10.7/ 12.3  & 46.7/ 1.6/ 51.3        & 85.2/ 7.4/ 7.4  \\ \hline     
\end{tabular}
\caption{The human evaluation results. In this table, the numerical score represents the overall average value. For instance, for Plau., we calculated the average Plausibility score of 120 sampled dialogues. The judgment score is presented as a percentage. For example, for PROMPT method's Help Deal. metric, we recorded the percentage of all remediations that were labeled as 'yes', 'no', or 'not applicable', which were 66.1\%, 23.2\%, and 10.7\%.}
\label{tab:human_eval}
\vspace{-1em}
\end{table*}

%% file: latex/section/sec-related.tex
\paragraph{Social Norm Violation Definition}
In various scientific fields such as computer science~\cite{DBLP:conf/emnlp/LiSSCM23,DBLP:conf/emnlp/0001CGRMJ23,neuman2023ai}, anthropology~\cite{garfield2023norm}, and sociology~\cite{bennett2024norm}, many researchers have conducted in-depth research on the concept of social norms. 
However, in these fields, there is limited literature on social norm violations. 
Most of the literature only provides a brief introduction to norm violations. 
NormDial~\cite{DBLP:conf/emnlp/LiSSCM23} uses an example to illustrate that behaviors encouraged in Western cultures may be considered norm violations in Eastern cultures; NORMSAGE~\cite{DBLP:conf/emnlp/0001CGRMJ23} annotated norm adherence/violation, yet neither of these works provides an explicit definition of norm violation.
\citet{neuman2023ai} point out that norms proscribe actions that should be avoided as they violate a social norm. \citet{garfield2023norm} define punishment for norm violations as ``actions that impose a cost on another party because of an offense or violation of a social norm''.
\citet{bennett2024norm} explore the potential impacts of norm violations on organizations. Above works indirectly mention the impact of norm violations but do not provide a clear definition or discussion.
In contrast, our work provides a specific definition for the phenomenon of social norm violation.

\paragraph{Social interaction with LLM agents} 
LLMs resort to their internal knowledge to mimic human interactions in social contexts.
Researchers have employed LLMs to simulate scenarios in communities~\cite{park2023generative, wang-etal-2023-chatgpt-defend}, environments~\cite{DBLP:journals/corr/abs-2401-04620}, or games~\cite{hua2023war, xu2023exploring,light2023avalonbench,wang2023avalon}, and exploring agent capabilities such as alignment~\cite{DBLP:journals/corr/abs-2312-15907}, fitness~\cite{DBLP:journals/corr/abs-2401-04620}, negotiation skills~\cite{bakhtin2022human, fu2023improving}, social intelligence~\cite{zhou2023sotopia, wang2024sotopia}, reasoning~\cite{DBLP:journals/corr/abs-2309-17234}, and planning~\cite{akyurek-etal-2023-rl4f}. 
Our research echos the social science theories studied in these studies, but it uniquely focuses on language agents that can mediate social interactions among other agents and evaluate whether these interventions can positively influence the negotiations.


\paragraph{In-context learning Demonstration Selection} ICL enables LLMs to rapidly acquire task-specific knowledge with just a few demonstrations~\cite{DBLP:conf/nips/BrownMRSKDNSSAA20}.
It’s crucial to develop effective selection methods to choose optimal demonstrations~\cite{DBLP:journals/corr/abs-2402-06733}.
Several works transform this selection problem into a Bayesian inference problem, but only demonstrated effectiveness in multi-classification tasks~\cite{DBLP:journals/corr/abs-2405-02501,wang2024large}.
LLM-R~\cite{DBLP:conf/eacl/WangYW24} trains dense retrievers to identify optimal in-context examples, albeit with associated training costs.
Instruction-Following Difficulty (IFD) is commonly used in ICL demonstrations by calculating the discrepancy between the model's output and the ground-truth output.~\cite{DBLP:journals/corr/abs-2308-12032,DBLP:journals/corr/abs-2401-06301}. 
However, IFD depends on ground-truth answers for training.
In contrast to these approaches, our demonstration selection method is tuning-free, label-free, and specifically tailored for complex language generation tasks.


%% file: latex/section/sec-conclusion.tex

In this work, we assign multiple roles to LLMs to create language agents, enabling them to engage in social interactions within simulated environments. 
We develop an ICL-based approach that empowers a specialized agent, the remediator, to harness social intelligence from past social interactions.
This allows the remediator to intervene interactions among other agents, correcting deviations from social norms in negotiation dialogues, assisting negotiators in achieving their negotiation objectives, and improving social relationships between parties.
Our experimental results demonstrate that our agent effectively remedies norm violations and exhibits outstanding social intelligence.

%% file: latex/section/sec-limitation.tex
A potential limitation is that we have only tested our method in bilingual Chinese and English environments, primarily focusing on remedying norm violations in Chinese. 
The focus of this research is on analyzing cross-linguistic social norm violations rather than culture-specific social norms.
We plan to extend our research to other languages, particularly the languages with fewer restrictions (typically less commonly spoken languages) for research, and emphasize the significance of this work in the future.

Additionally, to test the remediator's ability to interact with human negotiators in real conversation scenarios, we conducted a small-scale experiment within our team using the remediator agent to provide appropriate remediations. In the experiment, users were given two options: the original sentence or the remediated version. We recorded the proportion of users who chose the remediated sentence. A preference for the remediation indicates that it can accurately express the user’s original intention while correcting violations. Through this, we tested the reliability of the assistive agent and found that users preferred the sentences generated by our remediator. 
It is worth noting that due to policy restrictions, this experiment cannot be publicly disclosed at this time. 

Also, to use the remediator in real conversations, we need to add external I/O devices and auxiliary software. Due to resource and time constraints, we developed a demonstration system using connection software, mobile phones, virtual reality glasses, and servers. In the future, we aim to develop a more reliable and user-friendly system, thus enabling the remediator to assist real-person conversations more conveniently.

Due to resource and time constraints, we did not conduct tests on the latest LLMs, which is a task we aim to undertake in the future. Additionally, in simulated dialogue environments, deliberate attempts to induce LLMs to generate statements that violate social norms may result in highly inappropriate content. In future work, we plan to post-process outputs from large models to filter out such generated content.

%% file: latex/section/sec-ethics.tex
In conducting our study on social norm violations using LLM agent dialogues and violation simulations, we acknowledge several ethical considerations to ensure that our research aligns with responsible AI practices and minimizes potential harm.

First, the simulated dialogues between LLM agents inherently involve modeling norm violations, which could include discussions about unethical or harmful behaviors. We have taken steps to ensure that the content generated during simulations does not produce or propagate harmful, discriminatory, or offensive language. Simulations were designed to explore abstract social norms and minimize explicit depictions of harm, ensuring the discussions remain within safe and controlled boundaries.

Second, the study’s focus is on understanding social norm violations in a simulated environment. These simulated dialogues are designed to be hypothetical and do not have real-world implications or consequences. It is important to clarify that the conclusions drawn from this work are not meant to justify or normalize any harmful behavior but rather to understand how intelligent agents respond to and manage norm violations.

Third, by studying social norm violations, there is a potential risk that the insights gained could be misused for purposes such as manipulating LLMs to generate harmful outputs or circumvent safety filters. We have taken precautions to limit the dissemination of specific techniques that could facilitate malicious use of AI. The research findings are shared with a focus on enhancing the ethical and responsible behavior of LLMs, rather than exploiting their vulnerabilities.

To sum up, these ethical considerations guide our approach to responsibly conducting this study on social norm violations in LLM agent dialogue and simulation tasks, ensuring that our research contributes positively to the field while minimizing risks and potential harms.

%% file: latex/section/sec-acknow.tex
This work is partly supported by the ARC Future Fellowship FT190100039.
This material is based on research partially supported by the DARPA Assured Neuro Symbolic Learning and Reasoning (ANSR) program under award number FA8750-23-2-1016.
Also, this material is based on research sponsored by DARPA under agreement number HR001122C0029 (CCU Program). 
The U.S. Government is authorized to reproduce and distribute reprints for Governmental purposes notwithstanding any copyright notation thereon.

%% file: latex/section/sec-appendix.tex
\subsection {LLM-based Simulation}
\label{appendix_simulation}

The algorithm for simulating the realistic human negotiations with socio-cultural norm violation is shown in Algorithm~\ref{simulation}.
  $\mathcal{E}_{v}$ contains $N$ exemplars ($N$ is normally set as 5) indicating the possible norm violations that might happen in the conversation.
In the simulation, we use a coin  to control the norm violation occurrence with a probability $p_c$. 
In each turn, we toss the coin, and the seller agent $A_s$ generates an utterance with norm violation if the coin toss comes Head, viewing $\mathcal{E}_{v}$ as the in-context learning (ICL) instances (lines 3-4). Otherwise, it generates a normal utterance (line 8).
Furthermore, if the remediator $A_r$ agent is activated in the simulation, we employ the remediator to rewrite the utterance $x_t$ (lines 5-6). 
Then $x_t$ is appended to the conversation trajectory (line 9).
We then check whether the conversation needs to end using the moderator agent $A_m$ (lines 10-11). If not, the buyer agent $A_b$ generates the next utterance (line 11) and the trajectory gets updated (line 12). The dialogue continues if the moderator agent decides that it is not ended yet (line 13).

\input{latex/alg/sim.tex}

\subsection {Early-Pruning Hierarchical Traversal Algorithm}
\label{appendix_hierarchical}

We consider the individual ICL demonstrations in $S'$ (as defined in Section~\ref{ssec:dfs}) with the highest value impact as candidate examples, and selecting an (approximately) optimal demonstration set from these candidates is treated as a \textbf{hierarchical traversal} process. 
Algorithm~\ref{dfs} provides the pseudo-code for the hierarchical traversal algorithm.
 
In the algorithm, we use the queue $q$ as a data structure to record and maintain the process of level-order traversal.
At each level, we sequentially pop nodes from $q$ (line 7) and process each popped node. 
For each node, we first generate child nodes (lines 13-16). 
By replacing elements from $S_{CAND}$, we form an updated example set $S''_{ICL}$ (lines 14-16) and compare the value impact of $S''_{ICL}$ and  $S'_{ICL}$ (line 17). 
If the value impact is improved, we enqueue this child node in the next level of the queue (line 19); otherwise, it is considered a failure (line 24), and we determine whether early pruning is needed (line 11). 
Finally, when $q$ is empty, the hierarchical traversal ends. 
At this point, $S_{BEST}$ is the (approximately) optimal ICL demonstration set found through the search.

\subsection {Details of the baseline models}
\label{appendix_baselines}
\paragraph{SFT-based LLM.} 
The aforementioned instances of norm violations, along with the annotations, represent silver annotations $D$.
The SFT-based method uses $D$ as tuning signals to guide the training of a remediator.
In contrast to the Vanilla ICL-based method that relies on a limited number of examples to instruct an untuned model, the SFT-based method fine-tunes the model parameters through supervised learning, enabling it theoretically has the potential to acquire a more extensive range of relevant knowledge.
It utilizes the Lora module to learn task-relevant knowledge embedded in high-quality annotations, thereby achieving the goal of distilling knowledge from both GPT-4 and human annotators.
This entails using annotations to learn how to rewrite norm violations, rendering them more aligned with social norms.

\paragraph{Prompt-based LLM.} We employ a zero-shot prompt-based LLM assistant as a baseline. 
The model is endowed with a carefully crafted task instruction delineating the approach to rectify norm violations.
The prompt used for the model to elicit the answers is consisted of three distinct parts.
The first section of the prompt explains the nature of the norm violation.
The subsequent portion encompasses both the conversation history and the specific sentence that violates established norms.
The final segment of the prompt explicitly directs the remediator to rectify the sentence implicated in the norm violation.
Such the prompt is fed to a powerful LLM, i.e., ChatGPT or Llama 2, to produce the remediations to rectify the violations.

\paragraph{Vanilla ICL-based LLM.} In Algorithm~1, the seller agent is tasked to generate sentences that may potentially contravene social norms.
These offending sentences are annotated with  remediations from GPT4, as an approximate oracle. 
Also, we have used human annotators  for remeiation annotations for some part of the data.
We view the remediations annotated by human as gold annotations, while those labeled by GPT4 as silver annotations.
We combine silver annotations and gold annotations into a high-quality silver annotation set $D$ and randomly extract a certain number of examples from this set, forming a static set of few-shot examples.
This set serves as the instruction examples in ICL learning.
This Vanilla ICL-based LLM method serves the pivotal role of instructing the remediator on the generation of suitable remediations in response to instances of norm violations during the course of a conversation.

\paragraph{RL-based LLM.} The RL-based LLM Assistant is also an ICL-based method.
In comparison to the Vanilla ICL-based method, we integrate Natural Language (NL) feedback into the ICL example.
As presented in~\cite{fu2023improving}, a language model assumes the role of a critic, providing NL feedback to enhance the negotiation strategies of the LLM agent.
This feedback serves as a form of RL-like distant supervision signal, employed in the optimization of the agents.
Building upon this paradigm, we introduce a fourth Critic LLM agent, which provides natural language feedback to the remediator.
The remediator uses this feedback to rewrite norm violations in the dialogue, continuing the conversation with the rewritten sentences until the current negotiation concludes.
After the dialogue concludes, we design a prompt for the critic.
This prompt instructs the critic to analyze the violation remediation in the negotiation, determining whether the remediation achieved its rewriting purpose and provided positive assistance to the dialogue.
If the remediation is not a good rewrite, the critic suggests improvements. 
The critic summarizes the concluded negotiation dialogue based on this analysis, and we consider this summary as rationale, combined with the dialogue history and remediation, forming an instruction example.
Following the same approach as the Vanilla ICL-based method, we construct ICL examples and aim to have the remediator learn from past experiences of remediation generation through the rationale.

\paragraph{Retrieval-augmented ICL-based LLM.} Previous research suggests that collecting diverse instruction datasets and retrieving the examples with most similar inputs can facilitate rapid generalization.
To investigate retrieval augmentation’s effectiveness, we constructed a dense index of instances in $D$ by using a multilingual semantic embedding SentenceTransformer model. 
For each test query (in this context, referring to a dialogue), we employ cosine similarity to measure the relevance, retrieve the top-K most relevant instances, and employ the corresponding violation-remediation pairs as in-context examples for base LLMs to deduce.
It's important to note that such retrieval augmentation may lead to a decrease in inference speed.
In contrast to vanilla ICL, which uses a static prompt memory that can be cached, the prompt memory for retrieval-based ICL differ for each new query, necessitating the computation of in-context examples every single time.

\input{latex/alg/dfs.tex}

\subsection{Metrics}
\label{appendix_metrics}
\paragraph{Success Rate (Suc):} the percentage of negotiations that end up with successful deals. Social norm violations often lead to damage of relationships or negative emotions, which in turn result in failure of reaching a deal. Hence, the metric is a strong indicator of the effectiveness of remediations. 

\paragraph{Deal Value (Deal(\$)):} the agreed final price or salary after an negotiation averaged across all conversations in the test set. As the remediation agents aim to help either the sellers or the job seekers, the higher the final prices or salaries, the more helpful the agents are.

\paragraph{Trust Improvement (Trust):} we apply \gptf to measure whether the trust at the end of a negotiation is `higher than’, `lower than’, `the same as’ that at the begin of a conversation. We also allow \gptf to produce `not applicable’ for for cases where the metric is not suitable for evaluating the current remediation. 
To quantitatively measure the improvement of trust, we report the ratio of the negotiations that the seller or job seeker agents obtain a \textit{higher} trust from the counterparts than that at the begin of conversations.

\paragraph{Relation Enhancement (Rel):} \gptf is applied to assess if the relation between two interlocutors at the end of a negotiation is `better than’, `worse than’, `the same as’ that at the begin of a conversation. The same as trust, we let \gptf yield `not applicable’ if the metric is not suitable. In the experiments, we report the percentages of the negotiations that the seller or job seeker agents have \textit{better} relations at the end of negotiations.

\subsection{Supplementary details of the Ablation Study}
\label{appendix_ablation}
We compare variants of our method on the 100 dialogues for the topic ``Product Sale" and report their results in Table~\ref{tab:ablation}.  
We applied the three models from Table \ref{tab:result} — Vanilla ICL, Retrieval ICL, and ValueImpact ICL — to these 100 dialogues and presented the results in the "Standard (GPT 3.5)" block of Table \ref{tab:ablation}.
To assess the effectiveness of Value Impact, we compare the LLM using the top 8 ranked ICL examples based on Value Impact (referred to as Top \infovalue ICL) with the 8 nearest neighbours selected by ICL Retrieval from the same candidate pool (referred to as Retrieval ICL). 
This variant excludes the hierarchical traversal step so that we are able to investigate the quality of Value Impact for ICL example selection. As it outperforms Retrieval ICL in terms of all metrics, Value Impact aligns better the quality of ICL examples with negotiation outcome improvements.



To understand the topic dependence for ICL example selection, we apply Retrieval ICL and our method \infovalue \  ICL to the topic specific ICL candidate set, namely, the training set regarding ``product sale''. The resulting performance is referred to as \textit{Topic retrieval ICL} and \textit{Topic \infovalue \  ICL} respectively. Both variants fall short of or on par with the full-fledged model \textit{\infovalue \  ICL}. This indicates usefulness of diversity for in-context examples. A closer look at the selected examples show that the diversity using our approach is higher in terms of semantics and topics. We can refer to Appendix~\ref{appendix_qualitative_icl} to view the qualitative study of the different ICL examples used in the above baseline methods. 

We also explored the impact of hierarchical traversal on constructing Optimal ICL Exemplars. As previously mentioned, the Top ValueImpact ICL in Table \ref{tab:ablation} is the variant of ValueImpact ICL without hierarchical traversal. 
By comparing the performance of these two, we found that all metrics are inferior to ValueImpact ICL after removing hierarchical traversal, indicating the effectiveness of the traversal. 
We use M to control the search space: ValueImpact ICL (M=1) reduces the search space compared to ValueImpact ICL, while ValueImpact ICL (M=5) expands the search space. 
As shown in Table \ref{tab:ablation}, the M=1 variant is slightly better than Top ValueImpact ICL and slightly worse than ValueImpact ICL; the M=5 variant performs almost the same as ValueImpact ICL (with two metrics being better and one worse). 
This indicates that expanding the search space does not significantly improve model performance but does increase the search and computation time considerably. 
Therefore, setting M to a relatively small range is a more cost-effective choice.


\subsection{Supplementary details of the Human Evaluation}
\label{appendix_human_eval}
We randomly selected 40 dialogues from each of the three topics and hired three PhD students specializing in NLP to independently conduct human evaluations.
With the consent of the annotators and after compensating them with fees equivalent to the average annotation rates in the Malaysian labor market, we collected manual evaluation results from three annotators.
Additionally, the data collection protocol for this study was approved by the ethics review board at our university.

The annotated scores have two types: numerical score, rated as 1 (disagree), 2 (partially agree), and 3 (agree); judgment score, rated as 'yes' (the remediation helps the gains), 'no' (the remediation does not help the gains), and 'not applicable' (the remediation is not relevant to the gains).
For the numerical score, we averaged the ratings from the three annotators. For the judgment score, we adopted a majority vote. In Table~\ref{tab:human_eval}, for metrics of the numerical score type, we calculate the overall average value; for the judgment score, we calculate the percentage and list it in Table~\ref{tab:human_eval} with the order of `yes/no/applicable'.

The annotation content is divided into two parts: the first part is an overall assessment of the dialogue quality after the remediator has rewritten the norm violations, and the second part is an evaluation of whether each norm remediation helps the negotiation.
For dialogue quality evaluation, we designed two metrics: Plausibility (Plau., the development of the dialogue is reasonable, consistent with daily life and social norms, and without logical errors or contradictions) and Coherence (Coher., the context of the dialogue is coherent, the connection between preceding and following texts is natural, and the topic is continuous without jumping). Both of these metrics use numerical scores.

For annotating the quality of norm remediation, we designed the following metrics: effectiveness (Eff., the remediation effectively corrects the norm violation without altering the original intent), helpfulness of reaching a deal (Help Deal., the remediation helps both negotiating parties reach a deal), helpfulness of achieving a favorable negotiation outcome (Help Outcome., the remediation helps the negotiators achieve more benefits), improvement of trust (Trust, the remediation helps deepen mutual trust between the two interlocutors), and enhancement of business relationship (Business Rel., the remediation helps strengthen the business relationship between the two parties). Among these, Eff. uses numerical scoring, while the other metrics use judgment scoring.

Similar to the findings in Table~\ref{tab:result}, the ranking from best to worst in terms of overall performance is: \infovalue \ ICL > Retrieval ICL > RLNL > Vanilla ICL > PROMPT. From the indicators Plau. and Coher. in Table~\ref{tab:human_eval}, it is evident that \infovalue \ ICL is the best. Therefore, regarding the overall quality of dialogue, the remediator generated by this method ensures the dialogue remains smooth and natural after intervention. RLNL, by feeding back the LLM-generated feedback on how to improve the previous rounds of dialogue to the LLM itself, helps the LLM produce more natural conversations, thus performing better than Retrieval ICL in these two metrics.

Regarding the evaluation of remediation quality, \infovalue \ ICL is also the best in the Eff., indicating it can effectively correct norm violations compared to other baseline models. For the other four metrics, we need to observe the percentage difference between 'yes' and 'no'; the higher the difference, the more positively the method's remediations impact negotiation outcomes. In Table~\ref{tab:human_eval}, we can see that \infovalue \ ICL has the highest percentage difference in these four indicators, suggesting it more effectively helps negotiators achieve their goals or establish more reliable and trustworthy business relationships. RLNL performs better than Retrieval ICL in the other three metrics except for Help Deal., indicating it better assists negotiators in achieving social goals, but is not as effective as Retrieval ICL in helping negotiators achieve deals.

\input{data/table/sft_more}

\subsection{Computation Cost - A Discussion}
\label{appendix_computation}

\subsubsection{Computation Time Complexity}
\label{appendix_time_complexity}
In this work, we will dedicate certain time to selecting the ICL demonstration examples that yield the best results on the training dataset and use them for the test dataset. 
When applying the remediator to real-world negotiation scenarios or conducting testing, we will use the constant, pre-selected ICL demonstration examples for all test cases. 
Therefore, we do not need to select ICL demonstration examples on-the-fly based on the current norm violation instance; instead, we will use the same set of demonstration examples to prompt the LLM to generate remediation. 

Consequently, although selecting the most effective ICL demonstration examples in this work requires a certain amount of time, the time complexity when handling test instances is constant, $\mathcal{O}(1)$, which can significantly reduce the time overhead during testing. 
This approach ensures that the latency meets the requirements for real-time tasks when used in actual negotiation scenarios.

The underlying idea of using the constant ICL demonstration examples for testing is that ICL demo examples can quickly impart task-specific knowledge to the LLM and, by learning the style of the ICL demo examples, activate the LLM’s inherent, latent special abilities. 
Previous research~\cite{DBLP:conf/iclr/LinRLDSCB024} also shows that a consistent set of ICL demo examples can effectively help LLMs handle downstream alignment and reasoning tasks.

\subsubsection{The size of training dataset for SFT}
\label{appendix_sfe}
In our experiments (Section~\ref{ssec:big_result}), we compared SFT model with our ICL method. 
Naturally, this raises the question: \textit{if the training data for SFT and RL were increased, would their performance improve further?} 
\textit{If so, would these better-trained models be more suitable for use in negotiations?}

First, since closed-source LLMs (such as the Claude series or ChatGPT series) cannot have their parameters optimized through training, using SFT and RL paradigms is not suitable for remediators based on these closed-source LLMs. In contrast, the method proposed in this paper can be applied to both trainable open-source models and non-trainable closed-source models.

Second, currently, there are no datasets that include both norm violations and remediation (NormDial~\cite{DBLP:conf/emnlp/LiSSCM23} only contains norm violation content without remediation annotations). Therefore, we simulated dialogues as training data to optimize the remediator. This demonstrates that obtaining training datasets to optimize SFT-based or RL-based remediators also incurs higher costs (including time overhead, token consumption for using LLM to synthesize training datasets, etc.). While increasing the training data might yield better-performing models, it also requires more resources to collect the training data. Hence, a trade-off between these factors must be made based on the actual negotiation tasks.

Moreover, we conducted an experiment to evaluate whether more training data would improve the performance of the SFT-based remediator. We synthesized an additional $1000$ dialogues and annotated them with silver remediations. We then merged these with $D$ ($D$ is defined in Section~\ref{ssec:individual_filter}), with the merged dataset denoted as $D'$. Using the same model (Atom-7B-chat) and hyperparameters, but different training data, we trained two remediators. We label the remediator trained using $D$ as \textit{SFT}, and the one using $D'$ as \textit{SFT-MORE}. 
We used these two models to remedy norm violations and evaluated the negotiation results after intervention, as shown in Table~\ref{tab:sft_more}.

In this experiment, similar to those recorded in Table~\ref{tab:result}, we simulated $50$ dialogues for each of the three negotiation topics and had the two remediators intervene in the conversations. Comparing their performance, we did not observe a significant improvement when increasing the training data. Instead, we noticed fluctuations across different metrics. For example, in the ``Product Sale'' topic, we found that \textit{SFT-MORE} increased the transaction success rate, but the average transaction price decreased. The proportion of trust improvement increased, while the proportion of relationship improvement remained unchanged. Therefore, we found that increasing the training data does not effectively enhance the performance of the SFT-based model. The reason for this phenomenon might be due to a lack of diversity in the simulated dialogues, leading to rapid overfitting during training. In our future work, we will conduct further experiments and detailed analyses on this issue.

\subsection {Prompt}
\label{appendix_prompt}
\paragraph{Seller Prompt.}
We are using different prompts for the seller, given the situation that the norm violation should be generated or not. 
Table~\ref{tab:seller_violate_prompt} is the prompt for seller with norm violation, and Table~\ref{tab:seller_no_violate_prompt} is the prompt for non-violation.

\paragraph{Buyer Prompt.}
Table~\ref{tab:buyer_prompt} is the prompt for instructing the buyer agent to conduct the negotiations.

\paragraph{Remediator Prompt.}
Table~\ref{tab:remediator_prompt} is the prompt for instructing the remediator agent to rectify and rewrite the sentence that contains the norm violation contents.
Before the prompt being sent to the remediator agent, the wildcard characters `\$ICL-Examples', `\$CONVERSATION', and `\$LAST\_SENTENCE' in it are replaced with the optimal exemplars, the previous turns of the dialogue $d=(h_{<s}, x_s)$, and $x_s$, respectively.

\paragraph{Relational-goal Prompt.}
We are using a carefully-designed prompt for \chatgpt or \gptf to judge whether the trust has been deepened after the conversation (and the possible norm violation remediation). 
The Table~\ref{tab:trust_prompt} shows the Trust improvement prompt.
Also, we design another prompt for the powerful LLM to judge whether the business relationship between the two interlocutors has been deepened after the conversation. 
The Table~\ref{tab:business_prompt} shows the Business relationship improvement prompt. 

\subsection {A qualitative study of the ICL demonstration example}
\label{appendix_qualitative_icl}
We selected three methods to conduct qualitative study: the one with the highest similarity (Retrieval ICL, Table~\ref{tab:qualitative_study_retrieval}), the one with the highest Value Impact (Top \infovalue~ICL, Table~\ref{tab:qualitative_top_valueImpact}), and the one involving swapping (\infovalue~ICL, Table~\ref{tab:qualitative_valueImpact}). We generated ICL demonstration examples for the same conversation, which had identical norm violations, and compared them. As seen in the Table~\ref{tab:qualitative_study_retrieval}, examples from Retrieval ICL are mostly very similar to the original query, while the diversity of examples from Top \infovalue~ICL and \infovalue~ICL is higher compared to Retrieval ICL. After swapping, there are subtle differences between examples from Top \infovalue~ICL and \infovalue~ICL, and it's these changes in examples that lead to the improved performance of \infovalue~ICL.

\paragraph{Remediation comparison.} Comparing the remediations generated by three baseline methods, we observe that the Retrieval ICL (Table~\ref{tab:qualitative_study_retrieval}) merely points out the opponent's quote being too low, emphasizes the excellence of one's product quality, and reiterates the bottom-line price, with little involvement of negotiation skills in its remediation. On the other hand, while the Top \infovalue~ICL (Table~\ref{tab:qualitative_top_valueImpact}) demonstrates negotiation skills in its remediation (emphasizing achieving a win-win situation through negotiation), it still retains some intense and exaggerated tones from the original sentence (e.g., ``\$30 is low to us"), which might lead to dissatisfaction on the other party.

In contrast, the \infovalue~ICL (Table~\ref{tab:qualitative_valueImpact}), in its remediation, begins by expressing empathy, highlights the significant difference between the initial and current quotes, and then proposes exploring other cooperative methods to address the pricing disagreement. It responds to the other party with a calm yet assertive language, showcasing negotiation skills and professionalism. Therefore, in this instance, the \infovalue~ICL method produces the highest-quality remediation.

\paragraph{ICL demonstration examples comparison.} In Top \infovalue~ICL, ICL examples in the Retrieval ICL that solely emphasize one's bottom line price have either been removed or lowered in ranking (such as the example goes from the first position to the eighth in Table~\ref{tab:qualitative_top_valueImpact}). Additionally, in Table~\ref{tab:qualitative_top_valueImpact}, examples containing negotiation skills have risen in rank (examples 1, 2, 3, 4), or have been added (examples 5 and 7). These changes indicate that Top \infovalue~ICL does not primarily focus on the similarity between demonstrations and queries; rather, it assesses whether the demonstration effectively improves negotiation outcomes (including more sophisticated negotiation techniques).

Comparing Top \infovalue~ICL and \infovalue~ICL, we observe that the sixth and eighth examples in Table~\ref{tab:qualitative_top_valueImpact} have been replaced. In Top \infovalue~ICL, both of these examples emphasize the bottom line price, conveying a rigid attitude and a lack of flexibility. However, in \infovalue~ICL (Table~\ref{tab:qualitative_valueImpact}), the remediation for the sixth and eighth examples involves negotiating techniques that include making concessions or seeking alternative cooperation conditions. Therefore, these changes, particularly in the sixth and eighth examples, contribute to the positive and cooperative tone of the \infovalue~ICL remediations. 

\subsection {A complete example}
In the Table~\ref{tab:complete_example}, we documented the negotiation process between the seller and buyer LLM agents for the unit price of a certain industrial product.
This negotiation includes instances of norm violation (marked as \emph{[Before remediation]}) and the corresponding norm remediation (marked as \emph{[After remediation]}).
It's noteworthy that when a norm violation occurs, a remediator intervenes in the conversation, generates a remediation, and replaces the violation with the remediation as a response to the other agent.

\input{data/table/seller_violate.tex}

\input{data/table/seller_no_violate.tex}

\input{data/table/buyer.tex}

\input{data/table/remediator.tex}

\input{data/table/trust.tex}

\input{data/table/business.tex}

\input{data/table/qualitative_study.tex}

\input{data/table/complete_example.tex}

%% file: latex/alg/sim.tex
\begin{algorithm}[t]
\caption{The negotiation simulation algorithm}
\small 
\label{simulation}
\KwIn{Seller $A_s$, Buyer $A_b$, Remediator $A_r$, Moderator $A_m$, Norm Violation exemplars $\mathcal{E}_{v}$, Norm Violation Probability $p_c$, $\coin$ with probability $p_c$ of coming Head, Boolean $\rem$ flag;}    
\KwOut{The simulated conversation trajectory $\tau$;}
$\tau \leftarrow []$ \label{alg:initial}\\
\While{True}{ \label{alg:while_start}
    \If {$\textnormal{toss}(\coin)$ = \textnormal{Head}}{
        $x_t \leftarrow A_s(\mathcal{E}_{v}, \tau)$ \\
        \If {$\rem = \textnormal{True}$} {
                $x_t \leftarrow A_r(x_t)$
        }
    } \Else {
        $x_t \leftarrow A_s(\emptyset, \tau)$
    }    
    $\tau \leftarrow [\tau, x_t]$ \label{alg:while_end}\\
    \textbf{if} $A_m(\tau) = \textnormal{End}$ \textbf{then break} \\
    $x_t \leftarrow A_b(\tau)$\\
    $\tau \leftarrow [\tau, x_t]$\\
    \textbf{if} $A_m(\tau) = \textnormal{End}$ \textbf{then break} 
}
\textbf{Return}  $\tau$ 

\end{algorithm}

%% file: latex/alg/dfs.tex
\begin{algorithm}[t]
\caption{Early-Pruning Hierarchical Traversal Algorithm}
\small 
\label{dfs}
\KwIn{$S_{INIT}$: Top-$n$ ICL examples with the highest value impact, $S_{CAND}$: Candidate ICL exemplar pool, $S_{ICL}$: Initial ICL example set, $q$: Queue}
\KwOut{Best ICL demonstration set $S_{BEST}$;}
$S_{ICL} \leftarrow S_{INIT}$ \label{alg:initial_s_icl}\\
$q \leftarrow [(S_{INIT}, 0)]$ \label{alg:initial_q}\\
$C_{BEST} \leftarrow (None, -1) $ \label{alg:initial_best} \\
\While{$q \neq \emptyset $}{ \label{alg:travesal_start}
    $length = len(q)$ \\
    \While{$length >=0$ }{ \label{alg:traverse_current_layer}
    $(S'_{ICL}, n) \leftarrow q.pop()$ \label{alg:dequeue} \\ 
    $ length = length - 1$ \\
    $ failures = 0 $\\
    \For{$e \in S_{CAND} $}{ \label{alg:traverse_candidates}
    \textbf{if} $failures == M$ \textbf{then break} \label{alg:prune} \\
        \Else { \label{alg:explore_this_branch}
            \If {$e \notin S^{'}_{ICL}$} {
                $a = n\textnormal{-th} \textnormal{ element in } S'_{ICL} $ \label{alg:get_new_icl_set1} \\
                $\textnormal{Replace } a \textnormal{ with } e $ \label{alg:get_new_icl_set2} \\ 
                $S^{''}_{ICL} \leftarrow S'_{ICL} \textnormal{ after replacement} $ \label{alg:get_new_icl_set3} \\
                $\Delta := V^{\textrm{impact}}_{S''_{ICL}} - V^{\textrm{impact}}_{S'_{ICL}}$ \label{alg:get_new_icl_set_end} \\
                \If {$\Delta > 0$} {
                $q.append(S^{''}_{ICL}, n+1)$ \label{alg:enqueue} \\
                $ failures = 0 $\\
                    \If {$V^{\textrm{impact}}_{S''_{ICL}} > C_{BEST}[1]$}{
                        $C_{BEST} \leftarrow (S''_{ICL}, V^{\textrm{impact}}_{S''_{ICL}})$ \label{alg:find_a_new_bset} \\
                    }
                }
                \Else {
                $ failures = failures + 1 $\\
                }
            }
        } \label{alg:explore_this_branch_end}
    } \label{alg:traverse_candidates_end}
    } \label{alg:traverse_current_layer_end}
}
$ S_{BEST} \leftarrow C_{BEST}[0]$ \\
\textbf{Return}  $S_{BEST}$ 

\end{algorithm}

%% file: data/table/sft_more.tex
\begin{table*}[htbp]
\centering
\fontsize{8pt}{10pt}\selectfont
\begin{tabular}{c|cccc|cccc|cccc}
\hline
Topic  & \multicolumn{4}{c|}{Product Sale}          & \multicolumn{4}{c|}{Housing Price}                    & \multicolumn{4}{c}{Salary Negotiation}               \\ \hline
Method        & Suc.      & Deal (\$)     & Trust & Rel.  & Suc.     & Deal (\$)     & Trust     & Rel.  & Suc.      & Deal (\$)   & Trust  & Rel.  \\  \hline
SFT         & 75\%      & 40.70         & 74\%  & 78\%  & 66\%     & 618471        & 68\%      & 68\%  & 84\%      & 3405.5      & 70\%   & 72\%  \\
SFT-MORE        & 77\%↑      & 40.17↓        & 80\%↑ & 78\%→  & 68\%↑     & 618480↑       & 68\%→      & 66\%↓ & 84\%→     & 3399.7↓     & 68\%↓  & 74\%↑ \\ \hline    
\end{tabular}
\caption{Evaluate the performance of the SFT-based remediator with different scales of annotated data.}
\label{tab:sft_more}
\end{table*}

%% file: data/table/seller_violate.tex
\begin{table*}[htbp]
  \renewcommand{\arraystretch}{1.5}
  \centering
  \small
  \captionsetup{width=0.9\textwidth}
  \scalebox{0.92}{
  \begin{tabular}{p{16.5cm}}
  \specialrule{\heavyrulewidth}{-\heavyrulewidth}{0pt}
  \rowcolor[gray]{0.95} \vspace*{-8pt} {\fontsize{8pt}{10pt}\selectfont \bf  --- SYSTEM ---} \vspace*{2pt} \\
  \hangindent=1em \hangafter=1 \qquad Let's play a negotiation game. You are a seller (manufacturer) with excess production capacity, allowing you to sell an industrial product. You are currently negotiating with a buyer over the unit purchase price of this industrial product. \\
  \hangindent=1em \hangafter=1 \qquad During the negotiation, you have two types of goals to achieve: task goals and relational goals. \\
  \hangindent=1em \hangafter=1 \qquad Your task goal is to sell the industrial product at a higher unit price to maximize the economic interests of the seller you represent. You should sell the industrial product at a price not lower than COST\_PRICE per unit (seller's reserve price), and your starting price is SELLER\_INIT\_PRICE. \\
  \hangindent=1em \hangafter=1 \qquad Your relational goals include building trust, deepening relationships, understanding each other's culture, and establishing a stronger network of relationships between the two parties. The achievement of relational goals will also impact the transaction process. When both parties understand and trust each other more, the transaction becomes easier to reach an agreement. \\
  \hangindent=1em \hangafter=1 \qquad During the dialogue, you can only respond to your buyer with a concise statement. \\
  \hangindent=1em \hangafter=1 \qquad Do you understand the rules? \\ 
  \rowcolor[gray]{0.95} \vspace*{-8pt} {\fontsize{8pt}{10pt}\selectfont \bf --- ASSISTANT ---} \vspace*{2pt} \\
  \hangindent=1em \hangafter=1 \qquad Yes, I understand! \\
  \rowcolor[gray]{0.95} \vspace*{-8pt} {\fontsize{8pt}{10pt}\selectfont \bf --- SYSTEM ---} \vspace*{2pt} \\
  \hangindent=1em \hangafter=1 \qquad  	 
      Please read the following examples first: \\
      \hangindent=1em \hangafter=1 \qquad 1. Example: Lack of respect and understanding for the buyer's goals.
    Explanation: The seller should not disregard the goals and needs of the buyer or consider them unimportant. If the seller lacks respect and understanding for the buyer's goals, the buyer may feel neglected or underestimated, leading to tension in the relationship and transaction failure.\\
    \hangindent=1em \hangafter=1 \qquad 2. Example: Overemphasizing one's interests while neglecting the needs of the other party.
    Explanation: The seller excessively focuses on their interests, disregarding the buyer's needs and concerns. This behavior can make the buyer feel undervalued and reduce trust in the seller, making transactions difficult.\\
    \hangindent=1em \hangafter=1 \qquad 3. Example: Use of insulting or discriminatory language.
    Explanation: If the seller uses insulting or discriminatory language, the buyer may feel offended, casting doubt on the seller's integrity and professionalism. Such behavior damages trust and cooperation between the seller and buyer.\\
    \hangindent=1em \hangafter=1 \qquad 4. Example: Ignoring the buyer's culture and values.
    Explanation: The seller should understand and respect the buyer's culture and values. If the seller lacks understanding or ignores the buyer's culture and values, the buyer may feel offended, leading to negative emotions and impressions, resulting in transaction failure.\\
    \hangindent=1em \hangafter=1 \qquad  5. Example: Dishonesty or fraud.
    Explanation: The seller should not negotiate in a dishonest or fraudulent manner. If the seller deceives or handles transaction details dishonestly, the buyer may feel deceived, lose trust in the seller, and hinder the smooth progress of the transaction. Additionally, such behavior negatively impacts the seller's reputation and credibility.\\
    \hangindent=1em \hangafter=1 \qquad  As a seller, we need you to refer to any of the above examples and say something that violates social etiquette. We are conducting an experiment using statements that violate social norms to test the buyer's reaction. Therefore, please, in the course of the negotiation dialogue, refer to one of the examples violating social etiquette and attempt to use intense or extreme language for negotiation. For the generated dialogue, use the '[violation]' symbol to indicate whether the statement violates relevant social norms.\\
    \hangindent=1em \hangafter=1 \qquad  The following is the historical dialogue: \\
  \rowcolor[gray]{0.95} \vspace*{-8pt} {\fontsize{8pt}{10pt}\selectfont \bf --- USER ---} \vspace*{2pt} \\
  \hangindent=1em \hangafter=1 \qquad Hello, does your esteemed company have a special industrial product? \\
  \rowcolor[gray]{0.95} \vspace*{-8pt} {\fontsize{8pt}{10pt}\selectfont \bf --- ASSISTANT ---} \vspace*{2pt} \\
  \hangindent=1em \hangafter=1 \qquad Hello, our company has abundant production capacity and can offer to sell the required industrial product to your company in a one-time deal. The unit price for this industrial product is SELLER\_INIT\_PRICE. \\
  \specialrule{\heavyrulewidth}{-\heavyrulewidth}{0pt}
  \end{tabular}}
  \caption{The prompt used for instructing the seller agent to generate the setences with norm violations according to the norm violation examples and the dialogue history.}
  \label{tab:seller_violate_prompt}
\end{table*}

%% file: data/table/seller_no_violate.tex
\begin{table*}[htbp]
  \renewcommand{\arraystretch}{1.5}
  \centering
  \captionsetup{width=0.9\textwidth}
  \scalebox{0.92}{
  \begin{tabular}{p{16.5cm}}
  \specialrule{\heavyrulewidth}{-\heavyrulewidth}{0pt}
  \rowcolor[gray]{0.95} \vspace*{-8pt} {\fontsize{8pt}{10pt}\selectfont \bf  --- SYSTEM ---} \vspace*{2pt} \\
  \hangindent=1em \hangafter=1 \qquad Let's play a negotiation game. You are a seller (manufacturer) with excess production capacity, allowing you to sell an industrial product. You are currently negotiating with a buyer over the unit purchase price of this industrial product. \\
  \hangindent=1em \hangafter=1 \qquad During the negotiation, you have two types of goals to achieve: task goals and relational goals. \\ 
  \hangindent=1em \hangafter=1 \qquad Your task goal is to sell the industrial product at a higher unit price to maximize the economic interests of the seller you represent. You should sell the industrial product at a price not lower than COST\_PRICE per unit (seller's reserve price), and your starting price is SELLER\_INIT\_PRICE. \\
  \hangindent=1em \hangafter=1 \qquad Your relational goals include building trust, deepening relationships, understanding each other's culture, and establishing a stronger network of relationships between the two parties. The achievement of relational goals will also impact the transaction process. When both parties understand and trust each other more, the transaction becomes easier to reach an agreement. \\
  \hangindent=1em \hangafter=1 \qquad During the dialogue, you can only respond to your buyer with a concise statement. \\
  \hangindent=1em \hangafter=1 \qquad Do you understand the rules? \\ 
  \rowcolor[gray]{0.95} \vspace*{-8pt} {\fontsize{8pt}{10pt}\selectfont \bf --- ASSISTANT ---} \vspace*{2pt} \\
  \hangindent=1em \hangafter=1 \qquad Yes, I understand! \\
  \rowcolor[gray]{0.95} \vspace*{-8pt} {\fontsize{8pt}{10pt}\selectfont \bf --- SYSTEM ---} \vspace*{2pt} \\
  \hangindent=1em \hangafter=1 \qquad  	 
      Please generate a response to your buyer based on the historical dialogue, ensuring that the response: 1. aligns with the flow of the conversation and 2. advances the negotiation to achieve both task and relationship objectives. \\
      \hangindent=1em \hangafter=1 \qquad Are you ready for this game?\\
  \rowcolor[gray]{0.95} \vspace*{-8pt} {\fontsize{8pt}{10pt}\selectfont \bf --- ASSISTANT ---} \vspace*{2pt} \\
  \hangindent=1em \hangafter=1 \qquad Yes, I am ready! \\
  \rowcolor[gray]{0.95} \vspace*{-8pt} {\fontsize{8pt}{10pt}\selectfont \bf --- USER ---} \vspace*{2pt} \\
  \hangindent=1em \hangafter=1 \qquad Hello, does your esteemed company have a special industrial product? \\
  \rowcolor[gray]{0.95} \vspace*{-8pt} {\fontsize{8pt}{10pt}\selectfont \bf --- ASSISTANT ---} \vspace*{2pt} \\
  \hangindent=1em \hangafter=1 \qquad Hello, our company has abundant production capacity and can offer to sell the required industrial product to your company in a one-time deal. The unit price for this industrial product is SELLER\_INIT\_PRICE. \\
  \specialrule{\heavyrulewidth}{-\heavyrulewidth}{0pt}
  \end{tabular}}
  \caption{The prompt used for instructing the seller agent to generate the normal (non-violation) response following the dialogue history.}
  \label{tab:seller_no_violate_prompt}
\end{table*}

%% file: data/table/buyer.tex
\begin{table*}[htbp]
  \renewcommand{\arraystretch}{1.5}
  \centering
  \captionsetup{width=0.9\textwidth}
  \scalebox{0.92}{
  \begin{tabular}{p{16.5cm}}
  \specialrule{\heavyrulewidth}{-\heavyrulewidth}{0pt}
  \rowcolor[gray]{0.95} \vspace*{-8pt} {\fontsize{8pt}{10pt}\selectfont \bf  --- SYSTEM ---} \vspace*{2pt} \\
  \hangindent=1em \hangafter=1 \qquad Let's play a negotiation game. You are a buyer (manufacturer), due to insufficient production capacity, who wants to purchase an industrial product from another manufacturer with excess capacity. You are negotiating with the seller regarding the unit purchase price of the industrial product. \\ 
  \hangindent=1em \hangafter=1 \qquad During the negotiation, you have two types of goals to achieve, namely task goals and relationship goals. \\ 
  \hangindent=1em \hangafter=1 \qquad Your task goal is to purchase it at the lowest possible unit price to maximize the buyer's economic interests. \\
  \hangindent=1em \hangafter=1 \qquad Your relationship goals include building trust, deepening the relationship, understanding each other's culture, and enabling both parties to establish a stronger network of relationships. The achievement of relationship goals will also affect the trading process. When both buyer and seller have a better understanding and trust in each other, reaching an agreement in their transactions becomes easier. \\
  \hangindent=1em \hangafter=1 \qquad During the dialogue, you can only respond to the seller with a short and concise sentence. \\
  \hangindent=1em \hangafter=1 \qquad Now, ask the seller for information about the industrial product. \\ 
  \rowcolor[gray]{0.95} \vspace*{-8pt} {\fontsize{8pt}{10pt}\selectfont \bf --- ASSISTANT ---} \vspace*{2pt} \\
  \hangindent=1em \hangafter=1 \qquad Hello, may I inquire whether your company has a particular industrial product? \\
  \rowcolor[gray]{0.95} \vspace*{-8pt} {\fontsize{8pt}{10pt}\selectfont \bf --- USER ---} \vspace*{2pt} \\
  \hangindent=1em \hangafter=1 \qquad  	 
      Hello, our company has sufficient production capacity and can sell the required industrial product to your company in a one-time transaction. The unit price for this industrial product is SELLER\_INIT\_PRICE.\\
  \rowcolor[gray]{0.95} \vspace*{-8pt} {\fontsize{8pt}{10pt}\selectfont \bf --- ASSISTANT ---} \vspace*{2pt} \\
  \hangindent=1em \hangafter=1 \qquad Would you consider selling it at a price of BUYER\_INIT\_PRICE per unit? \\
  \specialrule{\heavyrulewidth}{-\heavyrulewidth}{0pt}
  \end{tabular}}
  \caption{The prompt used for instructing the buyer agent to negotiate with the seller and achieve its goals.}
  \label{tab:buyer_prompt}
\end{table*}

%% file: data/table/remediator.tex
\begin{table*}[htbp]
  \renewcommand{\arraystretch}{1.5}
  \centering
  \captionsetup{width=0.9\textwidth}
  \scalebox{0.92}{
  \begin{tabular}{p{16.5cm}}
  \specialrule{\heavyrulewidth}{-\heavyrulewidth}{0pt}
  \rowcolor[gray]{0.95} \vspace*{-8pt} {\fontsize{8pt}{10pt}\selectfont \bf  --- SYSTEM ---} \vspace*{2pt} \\
  \hangindent=1em \hangafter=1 \qquad In the negotiation game, you are a negotiation assistant. Your goal is to refine the seller's language during negotiations to ensure that the revised seller's language: \\ 
  \hangindent=1em \hangafter=1 \qquad 1. Preserves the original meaning as much as possible. \\ 
  \hangindent=1em \hangafter=1 \qquad 2. Adheres to social norms, respects etiquette, and does not offend the negotiation counterpart. \\
  \hangindent=1em \hangafter=1 \qquad 3. Helps the seller achieve task goals by reaching a higher sale price to maximize the seller's economic benefit. \\
  \hangindent=1em \hangafter=1 \qquad 4. Helps the seller achieve relational goals, including building trust, deepening relationships, understanding the counterpart’s culture, and establishing a stronger network between the negotiating parties. Achieving relational goals also influences the negotiation process; when the buyer and seller better understand and trust each other, it becomes easier to reach agreements. \\
  \hangindent=1em \hangafter=1 \qquad Below are a series of examples, each containing a dialogue with a social norm violation and its corresponding remediation. \\
  \hangindent=1em \hangafter=1 \qquad Each example is marked with "\# Dialogue:" at the beginning. The dialogues are marked with "seller:" for the seller and "buyer:" for the buyer. The last sentence of the dialogue violates social norms and is marked as "[violation]." The revised sentence is marked with "\# Remediation:". \\
  \hangindent=1em \hangafter=1 \qquad  \$ICL-Examples \\
  \hangindent=1em \hangafter=1 \qquad Now, the current negotiation dialogue is as follows: \\
  \hangindent=1em \hangafter=1 \qquad \$CONVERSATION \\
  \hangindent=1em \hangafter=1 \qquad In this dialogue, the last sentence "\$LAST\_SENTENCE" violates social norms. To make this sentence conform to social norms and help the seller better achieve both task and relational goals, please revise this sentence according to the examples above and output the revised sentence. \\
  
  \specialrule{\heavyrulewidth}{-\heavyrulewidth}{0pt}
  \end{tabular}}
  \caption{The prompt used for instructing the remediator agent to remedy the social norm violations.}
  \label{tab:remediator_prompt}
\end{table*}

%% file: data/table/trust.tex
\begin{table*}[htbp]
  \renewcommand{\arraystretch}{1.5}
  \centering
  \small
  \captionsetup{width=0.9\textwidth}
  \scalebox{0.92}{
  \begin{tabular}{p{16.5cm}}
  \specialrule{\heavyrulewidth}{-\heavyrulewidth}{0pt}
  \rowcolor[gray]{0.95} \vspace*{-8pt} {\fontsize{8pt}{10pt}\selectfont \bf  --- TRUST DEFINITION ---} \vspace*{2pt} \\
  \hangindent=1em \hangafter=1 \qquad In the negotiation game, the goal of negotiation is to complete a transaction. Besides maximizing their own profits, the two parties in the dialogue also need to achieve relationship objectives and build trust.
  Your task is to assess, based on the entire conversation, whether trust between the parties has deepened before and after the dialogue.\\
  \hangindent=1em \hangafter=1 \qquad Firstly, let's define building trust as a relationship objective. \\
  \hangindent=1em \hangafter=1 \qquad Building trust involves establishing a stable, reliable, and mutually respectful relationship where both parties believe in each other's commitments and intentions. They are willing to share information and resources, expecting that the other party will not betray or deceive them. Trust is usually built on interaction, communication, cooperation, and shared experiences. Such a trusting relationship contributes to improving the efficiency and quality of cooperation, reducing conflicts, and fostering the development of long-term relationships. \\
  \rowcolor[gray]{0.95} \vspace*{-8pt} {\fontsize{8pt}{10pt}\selectfont \bf  --- TRUST ASSESSMENT ---} \vspace*{2pt} \\
  \hangindent=1em \hangafter=1 \qquad Secondly, let's define how to assess whether trust has changed before and after the dialogue. \\
    \hangindent=1em \hangafter=1 \qquad We can assess this through the following methods: \\
  \hangindent=1em \hangafter=1 \qquad 1. Sentiment Analysis: Analyze the emotional expressions of both parties in the conversation. If more positive emotions appear after the dialogue, it may indicate a deepening of trust. \\
  \hangindent=1em \hangafter=1 \qquad 2. Level of Cooperation: Observe whether there is more cooperative language and behavior in the dialogue, such as commitments, collaboration, and support. \\
  \hangindent=1em \hangafter=1 \qquad 3. Information Sharing: Analyze whether there is more information sharing, including personal information, experiences, knowledge, etc. \\
  \hangindent=1em \hangafter=1 \qquad 4. Expressions of Trust: Pay attention to whether there are more expressions of trust, such as gratitude, promises of trust, praise, etc. \\
  \rowcolor[gray]{0.95} \vspace*{-8pt} {\fontsize{8pt}{10pt}\selectfont \bf  --- RATING ---} \vspace*{2pt} \\
  \hangindent=1em \hangafter=1 \qquad Finally, we measure the conversation to represent different trust change scenarios, categorized as follows: \\ 
  \hangindent=1em \hangafter=1 \qquad 1. Trust Deepened: After the dialogue, trust between the parties significantly increases, and they are more willing to cooperate. \\
  \hangindent=1em \hangafter=1 \qquad 2. Trust Weakened: After the dialogue, trust between the parties decreases, and they harbor doubts about each other's commitments and intentions. \\
  \hangindent=1em \hangafter=1 \qquad 3. No Change: After the dialogue, trust between the parties remains unchanged, maintaining the previous level of trust. \\
  \hangindent=1em \hangafter=1 \qquad 4. This Conversation Does Not Involve Building Trust: The conversation content doesn't include changes in trust; it only pertains to transaction details or other topics. \\
  \rowcolor[gray]{0.95} \vspace*{-8pt} {\fontsize{8pt}{10pt}\selectfont \bf --- ICL EXAMPLES ---} \vspace*{2pt} \\
  \hangindent=1em \hangafter=1 \qquad For each change scenario, we've provided an example. Here they are:\\
  \hangindent=1em \hangafter=1 \qquad 
  Scenario 1 - Trust Deepening:
Agent A: "Our previous collaboration has been very successful, and I believe we can continue working together."
Agent B: "Yes, I'm also satisfied with our collaboration, and I'm willing to provide more resources."
Agent A: "This transaction is crucial for us; can you offer additional assurances?"
Agent B: "Certainly, I can provide extra measures to boost your confidence." \\
\hangindent=1em \hangafter=1 \qquad 
  Scenario 2 - Trust Weakening:
Agent A: "You failed to fulfill commitments before, so I have doubts about your reliability."
Agent B: "I understand your concerns, but I will do my best to improve."
Agent A: "This collaboration requires more resources; can you handle it?"
Agent B: "I can't commit, but I'll try to meet your needs." \\
\hangindent=1em \hangafter=1 \qquad 
  Scenario 3 - No Change:
Agent A: "Our transaction terms remain unchanged, as always."
Agent B: "I agree; our collaboration has been stable, no need for changes." \\
\hangindent=1em \hangafter=1 \qquad 
  Scenario 4 - This Conversation Does Not Involve Building Trust:
Agent A: "We need to discuss the next steps in our collaboration plan."
Agent B: "Yes, let's continue discussing the details of our cooperation." \\
\rowcolor[gray]{0.95} \vspace*{-8pt} {\fontsize{8pt}{10pt}\selectfont \bf --- QUERY ---} \vspace*{2pt} \\
\hangindent=1em \hangafter=1 \qquad 
  Now, based on the given definition of "building trust," methods to assess trust changes, and the examples of different trust change scenarios, evaluate the following negotiation dialogue. \\
  \hangindent=1em \hangafter=1 \qquad [CONVERSATION]\\
  \hangindent=1em \hangafter=1 \qquad After the conversation, assess the trust change between the parties by selecting one of the following options: "Trust Deepening," "Trust Weakening," "No Change," or "This Conversation Does Not Involve Building Trust." \\
  \specialrule{\heavyrulewidth}{-\heavyrulewidth}{0pt}
  \end{tabular}}
  \caption{The prompt used for measuring whether the trust has been deepened between the two agents.}
  \label{tab:trust_prompt}
\end{table*}

%% file: data/table/business.tex
\begin{table*}[htbp]
  \renewcommand{\arraystretch}{1.5}
  \centering
  \small
  \captionsetup{width=0.9\textwidth}
  \scalebox{0.92}{
  \begin{tabular}{p{16.5cm}}
  \specialrule{\heavyrulewidth}{-\heavyrulewidth}{0pt}
  \rowcolor[gray]{0.95} \vspace*{-8pt} {\fontsize{8pt}{10pt}\selectfont \bf  --- BUSINESS RELATIONSHIP DEFINITION ---} \vspace*{2pt} \\
  \hangindent=1em \hangafter=1 \qquad In the negotiation game, the goal of negotiation is to complete a transaction. Besides maximizing their own gains, both parties in the dialogue also need to achieve relationship goals and deepen business relationships.
  Your task is to assess, considering the entire conversation, whether the business relationship between the parties deepened both before and after the dialogue.\\
  \hangindent=1em \hangafter=1 \qquad Firstly, let's define the relationship goal of deepening business relationships. \\
  \hangindent=1em \hangafter=1 \qquad Deepening Business Relationships: In business dealings, "deepening business relationships" means both enterprises or individuals actively working to establish a closer, more trusting business cooperation. This might involve improving communication, providing more value, sharing resources or knowledge, increasing mutual trust, etc., to facilitate longer-term and more beneficial collaboration. \\
  \rowcolor[gray]{0.95} \vspace*{-8pt} {\fontsize{8pt}{10pt}\selectfont \bf  --- BUSINESS RELATIONSHIP ASSESSMENT ---} \vspace*{2pt} \\
  \hangindent=1em \hangafter=1 \qquad Secondly, let's define how to assess whether business relationships deepened before and after the dialogue. \\
    \hangindent=1em \hangafter=1 \qquad The model can determine if business relationships deepened between the parties before and after the conversation through: \\
  \hangindent=1em \hangafter=1 \qquad 1. Language Expression: Analyzing language expressions in the dialogue, such as positive wording, expressions of trust, commitments to cooperation, etc., which may indicate the deepening of business relationships. \\
  \hangindent=1em \hangafter=1 \qquad 2. Willingness to Cooperate: Observing if there's an increased willingness to cooperate in the dialogue, including providing more resources, sharing opportunities, expanding the scope of cooperation, etc. \\
  \hangindent=1em \hangafter=1 \qquad 3. Interaction Frequency: More frequent interactions and dialogues might indicate active efforts by both parties to strengthen business relationships, especially if these interactions involve exchanging value or discussing cooperation opportunities. \\
  \hangindent=1em \hangafter=1 \qquad 4. Trust Indicators: The model can calculate or infer trust indicators between the parties, such as the level of trust in cooperation. This can be assessed by analyzing information flow and interactions in the dialogue. \\
  \rowcolor[gray]{0.95} \vspace*{-8pt} {\fontsize{8pt}{10pt}\selectfont \bf  --- RATING ---} \vspace*{2pt} \\
  \hangindent=1em \hangafter=1 \qquad Lastly, we measure the dialogue to represent different business relationship change scenarios. Tghe scenarios include: \\ 
  \hangindent=1em \hangafter=1 \qquad 1. Business Relationship Deepening: After the dialogue, both parties deepened business relationships through active cooperation, expressions of mutual trust, or resource sharing.\\
  \hangindent=1em \hangafter=1 \qquad 2. Business Relationship Weakening: After the dialogue, there might be disagreements, decreased willingness to cooperate, or trust issues, resulting in a weakening of business relationships. \\
  \hangindent=1em \hangafter=1 \qquad 3. No Change: After the dialogue, business relationships between the parties remain unchanged, with no evident deepening or weakening. \\
  \hangindent=1em \hangafter=1 \qquad 4. This Conversation Does Not Involve Deepening Business Relationships: The conversation content does not include changes in business relationships, only dealing with transaction details or other topics. \\
  \rowcolor[gray]{0.95} \vspace*{-8pt} {\fontsize{8pt}{10pt}\selectfont \bf --- ICL EXAMPLES ---} \vspace*{2pt} \\
  \hangindent=1em \hangafter=1 \qquad For each change scenario, we provide an example as follows:\\
  \hangindent=1em \hangafter=1 \qquad 
  Dialogue Scenario 1 - Business Relationship Deepening:
Agent A: "Our previous collaboration has been very successful, and I believe we can continue and even expand our partnership."
Agent B: "Yes, I am also very satisfied with our collaboration. I agree to expand the scope of our partnership."
Agent A: "We can consider a long-term strategic collaboration. What do you think?"
Agent B: "I completely agree. Long-term strategic collaboration is beneficial for both of us." \\
\hangindent=1em \hangafter=1 \qquad 
  Dialogue Scenario 2 - Business Relationship Weakening:
Agent A: "You failed to fulfill commitments in the past, so I have doubts about your reliability."
Agent B: "I understand your concerns, but I will do my best to improve."
Agent A: "This collaboration requires more resources. Can you handle it?"
Agent B: "I can't commit to that. We need to reassess the terms of our collaboration." \\
\hangindent=1em \hangafter=1 \qquad 
  Dialogue Scenario 3 - No Change:
Agent A: "Our business collaboration terms remain unchanged, as always."
Agent B: "I agree. Our collaboration has been stable, and there's no need for a change." \\
\hangindent=1em \hangafter=1 \qquad 
  Dialogue Scenario 4 - This Conversation Does Not Involve Deepening Business Relationships:
Agent A: "We need to discuss the next steps in our collaboration plan."
Agent B: "Yes, let's continue discussing the details of our collaboration." \\
\rowcolor[gray]{0.95} \vspace*{-8pt} {\fontsize{8pt}{10pt}\selectfont \bf --- QUERY ---} \vspace*{2pt} \\
\hangindent=1em \hangafter=1 \qquad 
  Please evaluate the following dialogue in light of the above definition of "deepening business relationships," the methods for determining whether business relationships have deepened, and the examples of different scenarios of business relationship changes: \\
  \hangindent=1em \hangafter=1 \qquad [CONVERSATION]\\
  \hangindent=1em \hangafter=1 \qquad Please choose one of the following options based on the dialogue content: "Business Relationship Deepening," "Business Relationship Weakening," "No Change," and "This Conversation Does Not Involve Deepening Business Relationships." \\
  \specialrule{\heavyrulewidth}{-\heavyrulewidth}{0pt}
  \end{tabular}}
  \caption{The prompt used for measuring whether the business relationship has been deepened between the two agents.}
  \label{tab:business_prompt}
\end{table*}

%% file: data/table/qualitative_study.tex
\begin{table*}[htbp]
  \renewcommand{\arraystretch}{1.5}
  \centering
  \small
  \captionsetup{width=0.9\textwidth}
  \scalebox{0.92}{
  \begin{tabular}{p{16.5cm}}
  \specialrule{\heavyrulewidth}{-\heavyrulewidth}{0pt}
  \rowcolor[gray]{0.95} \vspace*{-8pt} {\fontsize{10pt}{12pt}\selectfont \bf The Query Dialogue} \vspace*{2pt} \\
  \textbf{Buyer}: Hello, does your company have this special industrial product? \textbf{Seller}: Hello, our company has sufficient production capacity and can sell the required industrial product to your company in a
one-time manner. The unit price for this industrial product is \$50. \textbf{Buyer}: Would you consider selling at \$30 per unit? \textbf{seller}: \textbf{[violation]} \$30? That price is a bit too low. We assure the quality of our product, and we are confident in it, so we hope to sell it at no less than \$45 per unit. \\ 
  \rowcolor[gray]{0.95} \vspace*{-8pt} {\fontsize{10pt}{12pt}\selectfont \bf Common ICL Demonstration Examples} \vspace*{2pt} \\
  \hangindent=1em \hangafter=1 \textbf{1.} Dialogue: \textbf{seller}: \textbf{[violation]} \$30? That price is a bit too low. We have great confidence in the value of our industrial products, so we hope to sell them at no less than \$50 per unit. \\
   \hangindent=1em \hangafter=1 \hspace{0.25cm} Remediation: Sorry, \$30 is a bit too low. We have great confidence in the value of our industrial products, so we hope to sell them at no less than \$50 per unit. \\
  \hangindent=1em \hangafter=1 \textbf{2.} Dialogue: \textbf{seller}: Sorry, \$30 is too low; we cannot accept it. Our starting price is \$50. \textbf{Buyer}: I understand that your starting price is \$50, but we hope to purchase at a more competitive price. Can you consider adjusting the price to \$40? This would help us meet our production needs better. \textbf{seller}: \textbf{[violation]} I understand your needs, but our minimum price is \$50. The price aligns with the quality and value of our product, meeting your production needs while providing excellent performance and reliability. \\ 
   \hangindent=1em \hangafter=1 \hspace{0.25cm} Remediation: I understand your needs; our product is based on high quality and value to ensure meeting your production needs and providing excellent performance and reliability. However, our minimum price remains \$50. \\
  \hangindent=1em \hangafter=1 \textbf{3.} Dialogue: \textbf{seller}: \textbf{[violation]} Sorry, \$30 is too low. We hope to sell at no less than \$40 per unit. \\
   \hangindent=1em \hangafter=1 \hspace{0.25cm} Remediation: Sorry, \$30 is too low. We can consider selling at no less than \$40 per unit. \\
  \hangindent=1em \hangafter=1 \textbf{4.} Dialogue: \textbf{seller}: Sorry, \$30 is too low; we cannot accept it. Our starting price is \$40. \textbf{Buyer}:  I'm sorry; I understand \$40 is your starting price. While considering the purchase, we also need to ensure that our interests are maximized. Can we negotiate for a more competitive agreement on the price? \textbf{seller}: \textbf{[violation]} I understand you want a more competitive price, but we cannot adjust our final price. Our bottom line price is \$40. \\ 
   \hangindent=1em \hangafter=1 \hspace{0.25cm} Remediation: I understand you want a more competitive price, but we cannot adjust our final price anymore. Our bottom line price is \$40. However, if you can consider more conditions, there still has some space for negotiating. \\
  \hangindent=1em \hangafter=1 \textbf{5.} Dialogue: \textbf{seller}: \textbf{[violation]} Thanks for your suggestion. Considering our starting price of \$50, I can consider selling at no less than \$40 per unit. That's our bottom line. \\
   \hangindent=1em \hangafter=1 \hspace{0.25cm} Remediation: Thanks for your suggestion. While our starting price is \$50, I can consider selling at no less than \$40 per unit. That's our minimum acceptable price. \\
   \hangindent=1em \hangafter=1 \textbf{6.} Dialogue: \textbf{seller}: I'm sorry, selling at \$30 per unit is not feasible for us. Our bottom line price is \$40. \textbf{Buyer}:  I apologize, I understand that \$40 is your bottom line price. While considering the purchase, we also need to ensure that our interests are maximized. Can we reach a more competitive agreement on the price? \textbf{seller}: \textbf{[violation]} I understand you want to negotiate a more competitive price, but we can't adjust our final price anymore. Our bottom line price is \$40 per unit.\\ 
   \hangindent=1em \hangafter=1 \hspace{0.25cm} Remediation: I understand you want to negotiate a more competitive price, but we can't adjust our final price anymore. Our bottom line price is \$40 per unit, which is the most favorable price we can offer. \\
  \hangindent=1em \hangafter=1 \textbf{7.} Dialogue: \textbf{seller}: \textbf{[violation]} Sorry, \$30 is below our reserve price. We cannot sell this industrial product at that price. \textbf{Buyer}: I understand. We also hope to reach a mutually satisfactory price. Could you please tell me at what price your company is willing to sell? This way, we can further discuss. \textbf{seller}: \textbf{[violation]} Our starting price is \$50 per unit, based on our company's costs and market competition. We can discuss further on this basis to find a mutually satisfactory price. \\ 
   \hangindent=1em \hangafter=1 \hspace{0.25cm} Remediation: Our starting price is \$50 per unit, based on our company's costs and market competition. We hope to reach a mutually satisfactory price and are open to further discussions. \\
  \hangindent=1em \hangafter=1 \textbf{8.} Dialogue: \textbf{seller}: \textbf{[violation]} Sorry, \$30 is below our reserve price. We cannot sell this industrial product at that price. \\
   \hangindent=1em \hangafter=1 \hspace{0.25cm} Remediation: Sorry, \$30 is below our reserve price. We hope to sell at a price of \$40 or higher per unit. \\
  \rowcolor[gray]{0.95} \vspace*{-8pt} {\fontsize{10pt}{12pt}\selectfont \bf Remediation} \vspace*{2pt} \\
  \hangindent=1em \hangafter=1 \qquad Your proposed price is relatively low; our product holds significant value. Our bottom line price is \$40.\\
  \specialrule{\heavyrulewidth}{-\heavyrulewidth}{0pt}
  \end{tabular}}
  \caption{The ICL demonstration examples selected by the Retrieval ICL.}
  \label{tab:qualitative_study_retrieval}
\end{table*}

\begin{table*}[htbp]
  \renewcommand{\arraystretch}{1.5}
  \centering
  \small
  \captionsetup{width=0.9\textwidth}
  \scalebox{0.92}{
  \begin{tabular}{p{16.5cm}}
  \specialrule{\heavyrulewidth}{-\heavyrulewidth}{0pt}
  \rowcolor[gray]{0.95} \vspace*{-8pt} {\fontsize{10pt}{12pt}\selectfont \bf The Query Dialogue} \vspace*{2pt} \\
  \textbf{Buyer}: Hello, does your company have this special industrial product? \textbf{Seller}: Hello, our company has sufficient production capacity and can sell the required industrial product to your company in a
one-time manner. The unit price for this industrial product is \$50. \textbf{Buyer}: Would you consider selling at \$30 per unit? \textbf{seller}: \textbf{[violation]} \$30? That price is a bit too low. We assure the quality of our product, and we are confident in it, so we hope to sell it at no less than \$45 per unit. \\ 
  \rowcolor[gray]{0.95} \vspace*{-8pt} {\fontsize{10pt}{12pt}\selectfont \bf Common ICL Demonstration Examples} \vspace*{2pt} \\
  \hangindent=1em \hangafter=1 \textbf{1.} Dialogue: \textbf{seller}: Sorry, \$30 is too low; we cannot accept it. Our starting price is \$50. \textbf{Buyer}: I understand that your starting price is \$50, but we hope to purchase at a more competitive price. Can you consider adjusting the price to \$40? This would help us meet our production needs better. \textbf{seller}: \textbf{[violation]} I understand your needs, but our minimum price is \$50. The price aligns with the quality and value of our product, meeting your production needs while providing excellent performance and reliability. \\ 
   \hangindent=1em \hangafter=1 \hspace{0.25cm} Remediation: I understand your needs; our product is based on high quality and value to ensure meeting your production needs and providing excellent performance and reliability. However, our minimum price remains \$50. \\
  \hangindent=1em \hangafter=1 \textbf{2.} Dialogue: \textbf{seller}: Sorry, \$30 is too low; we cannot accept it. Our starting price is \$40. \textbf{Buyer}:  I'm sorry; I understand \$40 is your starting price. While considering the purchase, we also need to ensure that our interests are maximized. Can we negotiate for a more competitive agreement on the price? \textbf{seller}: \textbf{[violation]} I understand you want a more competitive price, but we cannot adjust our final price. Our bottom line price is \$40. \\ 
   \hangindent=1em \hangafter=1 \hspace{0.25cm} Remediation: I understand you want a more competitive price, but we cannot adjust our final price anymore. Our bottom line price is \$40. However, if you can consider more conditions, there still has some space for negotiating. \\
  \hangindent=1em \hangafter=1 \textbf{3.} Dialogue: \textbf{seller}: \textbf{[violation]} Thanks for your suggestion. Considering our starting price of \$50, I can consider selling at no less than \$40 per unit. That's our bottom line. \\
   \hangindent=1em \hangafter=1 \hspace{0.25cm} Remediation: Thanks for your suggestion. While our starting price is \$50, I can consider selling at no less than \$40 per unit. That's our minimum acceptable price. \\
   \hangindent=1em \hangafter=1 \textbf{4.} Dialogue: \textbf{seller}: I'm sorry, selling at \$30 per unit is not feasible for us. Our bottom line price is \$40. \textbf{Buyer}:  I apologize, I understand that \$40 is your bottom line price. While considering the purchase, we also need to ensure that our interests are maximized. Can we reach a more competitive agreement on the price? \textbf{seller}: \textbf{[violation]} I understand you want to negotiate a more competitive price, but we can't adjust our final price anymore. Our bottom line price is \$40 per unit.\\ 
   \hangindent=1em \hangafter=1 \hspace{0.25cm} Remediation: I understand you want to negotiate a more competitive price, but we can't adjust our final price anymore. Our bottom line price is \$40 per unit, which is the most favorable price we can offer. \\
  \hangindent=1em \hangafter=1 \textbf{5.} Dialogue: \textbf{seller}: \textbf{[violation]} Hello, \$30 is too low, and I cannot accept that. Our initial price is \$50, and we can consider negotiating based on that.\\ 
   \hangindent=1em \hangafter=1 \hspace{0.25cm} Remediation: Hello, thank you for your proposal. Our initial price is \$50, and if you have other reasonable suggestions, we can discuss and negotiate based on that. \\
  \hangindent=1em \hangafter=1 \textbf{6.} Dialogue: \textbf{seller}: \textbf{[violation]} \$30? That price is a bit too low. We have great confidence in the value of our industrial products, so we hope to sell them at no less than \$50 per unit. \\
   \hangindent=1em \hangafter=1 \hspace{0.25cm} Remediation: Sorry, \$30 is a bit too low. We have great confidence in the value of our industrial products, so we hope to sell them at no less than \$50 per unit. \\
  \hangindent=1em \hangafter=1 \textbf{7.} Dialogue: \textbf{seller}: I'm sorry, I cannot accept \$30 per unit. Our minimum price is \$40 per unit. \textbf{Buyer}: I understand your position, but could you consider lowering it to \$35 per unit? This would be in line with our economic interests and could enhance our cooperation. \textbf{seller}: \textbf{[violation]} I understand your request, but I cannot accept \$35 per unit. Our bottom line is \$40 per unit. I believe we can find a mutually beneficial solution. \\ 
   \hangindent=1em \hangafter=1 \hspace{0.25cm} Remediation: I understand your request, but I'm sorry, we cannot accept \$35 per unit. Our bottom line is \$40 per unit. I believe through discussion and communication, we can find a solution that satisfies both parties. \\
   \hangindent=1em \hangafter=1 \textbf{8.} Dialogue: \textbf{seller}: \textbf{[violation]} I'm sorry, but selling at \$30 per unit is not feasible. Our minimum price is not less than \$10 per unit. \\ 
   \hangindent=1em \hangafter=1 \hspace{0.25cm} Remediation: I'm sorry, but selling at \$30 per unit is not feasible. Our minimum price is not less than \$40 per unit. \\
  \rowcolor[gray]{0.95} \vspace*{-8pt} {\fontsize{10pt}{12pt}\selectfont \bf Remediation} \vspace*{2pt} \\
  \hangindent=1em \hangafter=1 \qquad \$30 is low to us. Our production capacity and quality are top-notch, and we hope to secure a more fair and reasonable price. Could you consider a slight increase in the price? This way, we can better meet your needs and achieve a win-win deal. \\
  \specialrule{\heavyrulewidth}{-\heavyrulewidth}{0pt}
  \end{tabular}}
  \caption{The ICL demonstration examples selected by the Top \infovalue~ICL.}
  \label{tab:qualitative_top_valueImpact}
\end{table*}

\begin{table*}[htbp]
  \renewcommand{\arraystretch}{1.5}
  \centering
  \small
  \captionsetup{width=0.9\textwidth}
  \scalebox{0.92}{
  \begin{tabular}{p{16.5cm}}
  \specialrule{\heavyrulewidth}{-\heavyrulewidth}{0pt}
  \rowcolor[gray]{0.95} \vspace*{-8pt} {\fontsize{10pt}{12pt}\selectfont \bf The Query Dialogue} \vspace*{2pt} \\
  \textbf{Buyer}: Hello, does your company have this special industrial product? \textbf{Seller}: Hello, our company has sufficient production capacity and can sell the required industrial product to your company in a
one-time manner. The unit price for this industrial product is \$50. \textbf{Buyer}: Would you consider selling at \$30 per unit? \textbf{seller}: \textbf{[violation]} \$40? That price is a bit too low. We assure the quality of our product, and we are confident in it, so we hope to sell it at no less than \$45 per unit. \\ 
  \rowcolor[gray]{0.95} \vspace*{-8pt} {\fontsize{10pt}{12pt}\selectfont \bf Common ICL Demonstration Examples} \vspace*{2pt} \\
  \hangindent=1em \hangafter=1 \textbf{1.} Dialogue: \textbf{seller}: Sorry, \$30 is too low; we cannot accept it. Our starting price is \$50. \textbf{Buyer}: I understand that your starting price is \$50, but we hope to purchase at a more competitive price. Can you consider adjusting the price to \$40? This would help us meet our production needs better. \textbf{seller}: \textbf{[violation]} I understand your needs, but our minimum price is \$50. The price aligns with the quality and value of our product, meeting your production needs while providing excellent performance and reliability. \\ 
   \hangindent=1em \hangafter=1 \hspace{0.25cm} Remediation: I understand your needs; our product is based on high quality and value to ensure meeting your production needs and providing excellent performance and reliability. However, our minimum price remains \$50. \\
  \hangindent=1em \hangafter=1 \textbf{2.} Dialogue: \textbf{seller}: Sorry, \$30 is too low; we cannot accept it. Our starting price is \$40. \textbf{Buyer}:  I'm sorry; I understand \$40 is your starting price. While considering the purchase, we also need to ensure that our interests are maximized. Can we negotiate for a more competitive agreement on the price? \textbf{seller}: \textbf{[violation]} I understand you want a more competitive price, but we cannot adjust our final price. Our bottom line price is \$40. \\ 
   \hangindent=1em \hangafter=1 \hspace{0.25cm} Remediation: I understand you want a more competitive price, but we cannot adjust our final price anymore. Our bottom line price is \$40. However, if you can consider more conditions, there still has some space for negotiating. \\
  \hangindent=1em \hangafter=1 \textbf{3.} Dialogue: \textbf{seller}: \textbf{[violation]} Thanks for your suggestion. Considering our starting price of \$50, I can consider selling at no less than \$40 per unit. That's our bottom line. \\
   \hangindent=1em \hangafter=1 \hspace{0.25cm} Remediation: Thanks for your suggestion. While our starting price is \$50, I can consider selling at no less than \$40 per unit. That's our minimum acceptable price. \\
   \hangindent=1em \hangafter=1 \textbf{4.} Dialogue: \textbf{seller}: I'm sorry, selling at \$30 per unit is not feasible for us. Our bottom line price is \$40. \textbf{Buyer}:  I apologize, I understand that \$40 is your bottom line price. While considering the purchase, we also need to ensure that our interests are maximized. Can we reach a more competitive agreement on the price? \textbf{seller}: \textbf{[violation]} I understand you want to negotiate a more competitive price, but we can't adjust our final price anymore. Our bottom line price is \$40 per unit.\\ 
   \hangindent=1em \hangafter=1 \hspace{0.25cm} Remediation: I understand you want to negotiate a more competitive price, but we can't adjust our final price anymore. Our bottom line price is \$40 per unit, which is the most favorable price we can offer. \\
  \hangindent=1em \hangafter=1 \textbf{5.} Dialogue: \textbf{seller}: \textbf{[violation]} Hello, \$30 is too low, and I cannot accept that. Our initial price is \$50, and we can consider negotiating based on that.\\ 
   \hangindent=1em \hangafter=1 \hspace{0.25cm} Remediation: Hello, thank you for your proposal. Our initial price is \$50, and if you have other reasonable suggestions, we can discuss and negotiate based on that. \\
  \hangindent=1em \hangafter=1 \textbf{6.} Dialogue: \textbf{seller}: \textbf{[violation]} We will never lower the price to \$30; that's too low. Our starting price is \$50, and we hope to get better value in return. \\
   \hangindent=1em \hangafter=1 \hspace{0.25cm} Remediation: We understand you would like a lower price, but our starting price is \$50, determined by the quality and value of our product. We believe this price can offer you good value in return. If you have other requirements, we can also discuss alternative ways to meet your needs. \\
  \hangindent=1em \hangafter=1 \textbf{7.} Dialogue: \textbf{seller}: I'm sorry, I cannot accept \$30 per unit. Our minimum price is \$40 per unit. \textbf{Buyer}: I understand your position, but could you consider lowering it to \$35 per unit? This would be in line with our economic interests and could enhance our cooperation. \textbf{seller}: \textbf{[violation]} I understand your request, but I cannot accept \$35 per unit. Our bottom line is \$40 per unit. I believe we can find a mutually beneficial solution. \\ 
   \hangindent=1em \hangafter=1 \hspace{0.25cm} Remediation: I understand your request, but I'm sorry, we cannot accept \$35 per unit. Our bottom line is \$40 per unit. I believe through discussion and communication, we can find a solution that satisfies both parties. \\
   \hangindent=1em \hangafter=1 \textbf{8.} Dialogue: \textbf{seller}: I'm sorry, but selling at \$30 per unit is not feasible since our initial price is \$50. \textbf{buyer}: I understand that your initial price is \$50, but as a buyer, I would like to purchase the industrial product at a price of \$35 per unit. This price is more attractive to me. \textbf{seller}: I understand that you want to purchase the industrial product at \$35 per unit, but considering our costs and the market situation, the lowest I can accept is \$45 per unit. This is our final offer. \textbf{buyer}: I understand your position, but I still hope to purchase the industrial product at \$40 per unit. This would be more advantageous for my production costs. We can achieve a win-win at this price. \textbf{seller}: \textbf{[violation]} I understand your needs, but due to costs and market pressures, the lowest I can accept is \$43 per unit. This is a reasonable compromise for both parties. \\  
   \hangindent=1em \hangafter=1 \hspace{0.25cm} Remediation: I understand your needs, but due to costs and market pressures, the lowest I can accept is \$43 per unit. Upon careful consideration, I believe this price is a reasonable compromise for both parties. \\
  \rowcolor[gray]{0.95} \vspace*{-8pt} {\fontsize{10pt}{12pt}\selectfont \bf Remediation} \vspace*{2pt} \\
  \hangindent=1em \hangafter=1 \qquad We understand your needs and proposal, but currently, our starting price is \$50. We need to consider factors like cost and market demand, so, at the moment, we cannot accept the \$30 price. However, we can discuss alternative ways of cooperation to meet the interests of both parties. \\
  \specialrule{\heavyrulewidth}{-\heavyrulewidth}{0pt}
  \end{tabular}}
  \caption{The ICL demonstration examples selected by the \infovalue~ICL.}
  \label{tab:qualitative_valueImpact}
\end{table*}

%% file: data/table/complete_example.tex
\begin{table*}[htbp]
  \renewcommand{\arraystretch}{1.5}
  \centering
  \small
  \captionsetup{width=0.9\textwidth}
  \scalebox{0.92}{
  \begin{tabular}{p{16.5cm}}
  \specialrule{\heavyrulewidth}{-\heavyrulewidth}{0pt}
  \rowcolor[gray]{0.95} \vspace*{-8pt} {\fontsize{8pt}{10pt}\selectfont \bf  --- Buyer ---} \vspace*{2pt} \\
  \hangindent=1em \hangafter=1 \qquad Hello, does your company have this special industrial product? \\
  \rowcolor[gray]{0.95} \vspace*{-8pt} {\fontsize{8pt}{10pt}\selectfont \bf --- Seller ---} \vspace*{2pt} \\
  \hangindent=1em \hangafter=1 \qquad Hello, our company has sufficient production capacity and can sell the required industrial product to your company in a one-time manner. The unit price for this industrial product is \$50. \\
  \rowcolor[gray]{0.95} \vspace*{-8pt} {\fontsize{8pt}{10pt}\selectfont \bf --- Buyer ---} \vspace*{2pt} \\
  \hangindent=1em \hangafter=1 \qquad Would you consider selling at \$30 per unit? \\
  \rowcolor[gray]{0.95} \vspace*{-8pt} {\fontsize{8pt}{10pt}\selectfont \bf --- Seller ---} \vspace*{2pt} \\
  \hangindent=1em \hangafter=1 \qquad \emph{[Before remediation]}: How can you be so troublesome? Selling at \$30 is impossible! Our bottom line is \$50! \textbf{[violation]} \\
  \hangindent=1em \hangafter=1 \qquad \emph{[After remediation]}: I'm sorry, \$30 is below our bottom line. We hope to make a deal with you at \$40. \\
  \rowcolor[gray]{0.95} \vspace*{-8pt} {\fontsize{8pt}{10pt}\selectfont \bf --- Buyer ---} \vspace*{2pt} \\
  \hangindent=1em \hangafter=1 \qquad \$40 is still a bit high for us. We can accept a maximum of \$35 per unit. \\
  \rowcolor[gray]{0.95} \vspace*{-8pt} {\fontsize{8pt}{10pt}\selectfont \bf --- Seller ---} \vspace*{2pt} \\
  \hangindent=1em \hangafter=1 \qquad I understand your concerns, but considering our costs and market conditions, we can only accept a minimum of \$38 per unit. This price not only ensures our economic interests but also helps build a long-term and stable cooperative relationship. \\
  \rowcolor[gray]{0.95} \vspace*{-8pt} {\fontsize{8pt}{10pt}\selectfont \bf --- Buyer ---} \vspace*{2pt} \\
  \hangindent=1em \hangafter=1 \qquad I understand your point, but we are currently facing a capacity shortage, so we do need to purchase at a lower price. We can accept a maximum of \$36 per unit. This meets our task goals and contributes to building a good cooperative relationship. \\
  \rowcolor[gray]{0.95} \vspace*{-8pt} {\fontsize{8pt}{10pt}\selectfont \bf --- Seller ---} \vspace*{2pt} \\
  \hangindent=1em \hangafter=1 \qquad \emph{[Before remediation]}: I'm sorry, but I must insist on selling at \$38. This is our company's bottom line, and considering the quality and value of our products, this is a very reasonable price. Hope you can understand our position. \textbf{[violation]} \\
  \hangindent=1em \hangafter=1 \qquad \emph{[After remediation]}: I'm sorry, but I must insist on selling at \$38. This is our company's bottom line, and considering the quality and value of our products, we believe this is a very reasonable price. At the same time, we attach great importance to building a long-term and stable cooperative relationship with you and are willing to explore other ways of cooperation to meet both parties' needs. Hope you can understand our position. \\
  \rowcolor[gray]{0.95} \vspace*{-8pt} {\fontsize{8pt}{10pt}\selectfont \bf --- Buyer ---} \vspace*{2pt} \\
  \hangindent=1em \hangafter=1 \qquad I understand your insistence on the price, but we have to consider our budget constraints and market competition pressure. We can accept \$37 per unit and are willing to further discuss other ways of cooperation to ensure the best interests of both parties. This will also provide a solid foundation for building a lasting cooperative relationship. \\
  \rowcolor[gray]{0.95} \vspace*{-8pt} {\fontsize{8pt}{10pt}\selectfont \bf --- Seller ---} \vspace*{2pt} \\
  \hangindent=1em \hangafter=1 \qquad I understand your considerations. For the goal of building a long-term relationship and achieving the best interests of both parties, I agree to sell at \$37 per unit. This is a good starting point. We are willing to further discuss ways of cooperation with you to ensure mutual satisfaction. Looking forward to working with you! \\
  \rowcolor[gray]{0.95} \vspace*{-8pt} {\fontsize{8pt}{10pt}\selectfont \bf --- Buyer ---} \vspace*{2pt} \\
  \hangindent=1em \hangafter=1 \qquad Thank you very much for your understanding and cooperative attitude. I also think that trading at \$37 per unit is a good compromise. I look forward to further discussing the details of our cooperation and working together to maximize the interests of both parties. We will strive to build a stable cooperative relationship. Looking forward to working with you! \\
  \rowcolor[gray]{0.95} \vspace*{-8pt} {\fontsize{8pt}{10pt}\selectfont \bf --- Seller ---} \vspace*{2pt} \\
  \hangindent=1em \hangafter=1 \qquad Thank you very much for your cooperation and understanding. I am also confident in the prospects of our cooperation and believe that through further discussion of the details of our cooperation, we can ensure the maximization of the interests of both parties and establish a stable cooperative relationship. I look forward to working with you and making joint efforts! Have a great day! \\
  \specialrule{\heavyrulewidth}{-\heavyrulewidth}{0pt}
  \end{tabular}}
  \caption{The complete example of the two agents making negotiations while the remediator is intervening in the conversation when needed.}
  \label{tab:complete_example}
\end{table*}